\documentclass[a4paper]{article}

\usepackage{graphicx}
\usepackage{subfigure}
\usepackage[margin=20mm]{geometry}

\date{}

\begin{document}

\title{Image-based Portrait Engraving}

\author{Paul L. Rosin\thanks{e-mail: Paul.Rosin@cs.cf.ac.uk}\\School of Computer Science \& Informatics\\Cardiff University %
\and Yu-Kun Lai\thanks{e-mail: Yukun.Lai@cs.cf.ac.uk}\\School of Computer Science \& Informatics\\Cardiff University}

\maketitle

\begin{abstract}
This paper describes a simple image-based method that applies engraving stylisation to portraits using ordered dithering.
Face detection is used to estimate a rough proxy geometry of the head consisting
of a cylinder, which is used to warp the dither matrix, causing the engraving lines to curve around the face for better stylisation.
Finally, an application of the approach to colour engraving is demonstrated.
\end{abstract}

\section{Introduction}

Engraving is a long established technique for the reproduction of artworks, e.g. in books or as stand-alone artworks.
Although there are a variety of styles, engraving often consists of
long, thin, gently curving, and continuous lines, whose thickness may vary along their length.
Sets of equidistant, roughly parallel lines form hatching, while two sets of intersecting hatchings create cross-hatching.
The varying line thickness and position as well as the presence or absence of cross-hatching enables engraving to capture the tonal density of the original artwork.
Engravers often align the orientations of the curves so that they follow the structure of content of the illustration.
It is also possible to vary the strokes in various ways (e.g. using short lines).
All together this enables engraving to provides visual cues that indicate form, texture, and shading simultaneously.

The field of non-photorealistic rendering (NPR)~\cite{NPRbook} has been applied to many artistic styles, including hatching and engraving.
Having a 3D model as input facilitates generating the engraving lines, since
direct knowledge of the object/scene structure provides guidance for orienting the lines~\cite{leister1994computer,zander,praun2001real},
e.g. they can be aligned with the principal curvature directions.
If the input is restricted to a 2D image then generating an engraving is more challenging,
and so many image-based engraving algorithms require the user to provide additional information.
An early example is the Dig$^{i}_{D}${\"u}rer~\cite{pnueli1994dig} system which uses curve evolution based on solving the eikonal equation.
However, one part of their scheme to avoid unwanted discontinuities involves semi-automatic segmentation of the image to
force discontinuities in the solution to occur at region boundaries.
Subsequent image-based methods used
a direction field defining the desired orientation of the lines~\cite{salisbury1997orientable},
or a segmentation of the image into regions, each with parametric grids defining the local line direction~\cite{Ostromoukhov}.a
In addition to the above, NPR has also explored other related styles that construct images from binary lines, such as
paths formed from minimal spanning trees, travelling salesman tours, dendrite stylisation, mazes (see~\cite{kaplan2013depiction} for an overview),
particle systems~\cite{wei2014coordinated}, circular scribbles~\cite{chiu2015tone} and amplitude modulated sinusoidal curves~\cite{ahmed2016amplitude}.

Portraiture is considered an importance element of art;
the National Portrait Gallery in London alone contains 195,000 portraits.
Consequently, there is a substantial body of NPR work dealing specifically with portraits.
Examples of work from the early days are simple line drawings~\cite{li1995extraction,brennan2007caricature},
through to recent deep learning approaches~\cite{Selim,yaniv2019face}.
Along the way, many different styles have been achieved, such as
generic cartoons~\cite{zhang2017data},
Simpson cartoon style~\cite{lee2013simpson},
paper-cut~\cite{meng2010artistic},
watercolour~\cite{rosin2017watercolour},
artistic line drawings~\cite{YiLLR19},
Julian Opie~\cite{rosin-portrait},
psychedelic/painterly~\cite{dipaola2016using},
and other painterly styles~\cite{wang2013learnable,zhang2014exemplar},
as well as application to video~\cite{fivser2017example,wang2017adaptive}.

This paper describes a simple method that applies engraving stylisation to portrait images using ordered dithering.
A dither matrix is first created that generates horizontal hatched lines with orthogonal cross-hatchings.
A simple proxy geometry for the human head is estimated for the human head, based on facial landmarks detected in the input image.
Using this geometry the engraving lines are warped so that they curve around the face.
Finally, the method's application to create colour engraving is demonstrated.

\section{Ordered Dithering}
\label{dithering}

\begin{figure}[htb]
\centering
\includegraphics[height=20mm]{blank} \includegraphics[height=20mm]{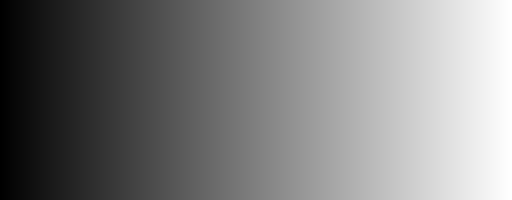}\\
\includegraphics[height=20mm]{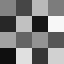} \includegraphics[height=20mm]{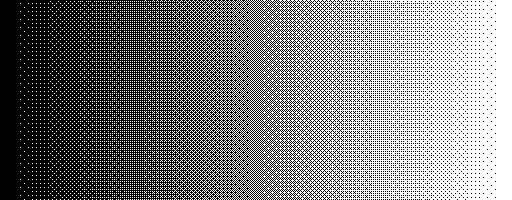}\\
\includegraphics[height=20mm]{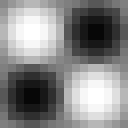} \includegraphics[height=20mm]{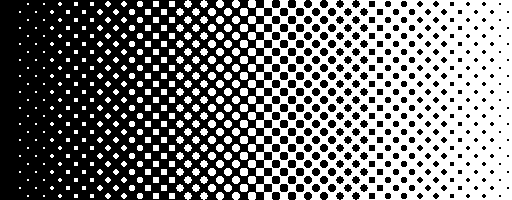}\\
\includegraphics[height=20mm]{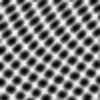} \includegraphics[height=20mm]{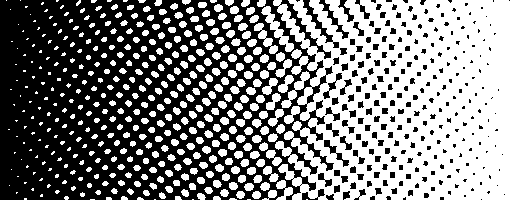}\\
\includegraphics[height=20mm]{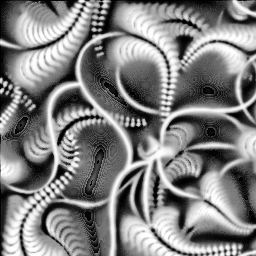} \includegraphics[height=20mm]{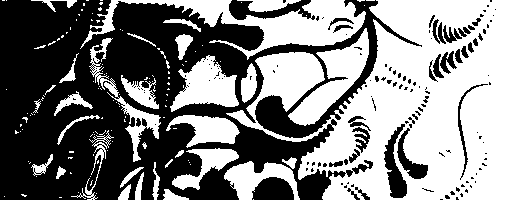}\\
\includegraphics[height=20mm]{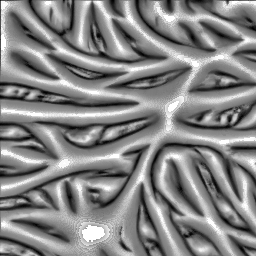} \includegraphics[height=20mm]{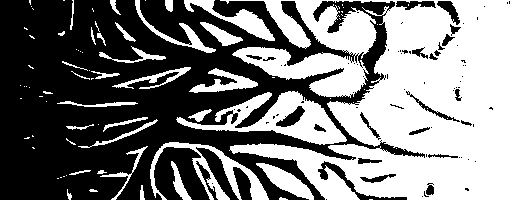}\\
\caption{Application of ordered dithering to an intensity ramp.
The left column shows the dither matrix, not to scale with dithered ramp on the right.
Except for the first two examples only a portion of each dither matrix is shown.
The first example uses a $4 \times 4$ Bayer matrix.}
\label{dither1}
\end{figure}

Ordered dithering is a simple, effective and efficient method for converting a greyscale image to black and white,
whilst locally preserving its average intensity.
It uses a ``dither matrix'' which tiles the intensity image, and defines a threshold for each pixel of that image.
Whereas techniques such as the Bayer matrix produce many isolated black and white pixels in the thresholded image
in order to minimise the visibility of the binary elements,
other dither matrices group adjacent values so as to create structures
(see examples in figure~\ref{dither1}), and this can be used for artistic effect~\cite{veryovka2000texture}.
So as to preserve the tonal distribution of the input image the values in the dither matrix should be uniformly distributed
over the tonal range (e.g. 0-255).

\begin{figure}[htbp]
\centering
\subfigure[]{\frame{\includegraphics[height=20mm]{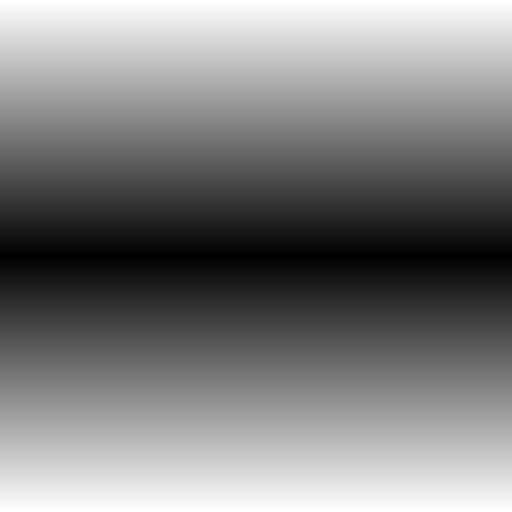}}}
\subfigure[]{\frame{\includegraphics[height=20mm]{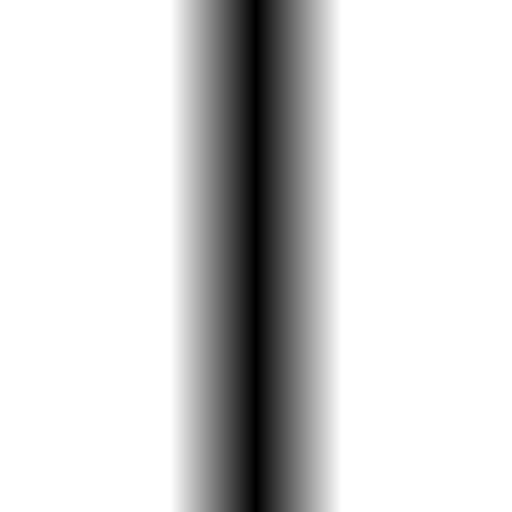}}}
\subfigure[]{\frame{\includegraphics[height=20mm]{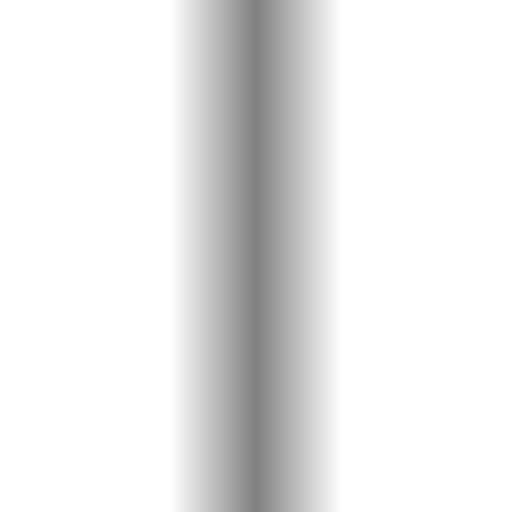}}}
\caption{Construction of the dither matrix combines
a) a dither matrix that produces horizontal lines, and
c) a scaled dither matrix that produces narrower vertical lines.}
\label{ditherConstruction}
\end{figure}

We wish to construct a dither matrix that will create hatching lines of varying thickness.
Ostromoukhov~\cite{Ostromoukhov} describes a scheme for constructing artistic dither matrices,
but we shall describe here a different intuitive approach that is sufficient for our purpose.
We first build the dither matrix shown in figure~\ref{ditherConstruction}a which will produce a horizontal line of appropriate thickness.
It is defined by rows of constant intensity, containing values $\{255, 254, 253, \ldots 1, 0, 1, \ldots 253, 254, 255\}$.
The cross-hatching will be perpendicular to these horizontal lines, but they need to have a different strength.
Otherwise the combination of equal thickness horizontal and vertical lines will produce a
grid effect containing spots instead of appearing primarily line-like.
Thus the vertical line for cross-hatching (figure~\ref{ditherConstruction}b)
is reduced in width by a half and the remainder of its dither matrix is saturated at peak value.
Furthermore, by scaling the dither matrix values to reduce their range
(i.e. intensity $x$ is scaled by $S$ as $f(x,S)=255 - S \times (255-x)$)
then the presence of cross-hatching can be restricted to only appear in the very dark and very light regions.
Note that values of $S$ should be in the range $[0,0.5]$
to ensure that the white vertical lines
only appear in the dark (i.e. darker than mid-gray)
regions, and vice-versa for the black vertical lines.

The two dither matrices
(for horizontal lines in figure~\ref{ditherConstruction}a and for vertical lines in figure~\ref{ditherConstruction}c)
are combined as follows.
The upper and lower thirds of the final dither matrix are constructed by
ANDing the two dither matrices, such that white vertical lines will be inserted into the thresholded result.
To insert black vertical lines a similar process is applied to this resulting dither matrix.
An inverted version of the vertical line dither matrix is
ORed with an inverted version of the dither matrix for vertical lines,
and this is only applied in the middle horizontal band
(the half area of the matrix that was not previously updated for white vertical lines).
Finally, the dither matrix is histogram equalised.

\begin{figure}[htbp]
\centering
(a) \includegraphics[height=20mm]{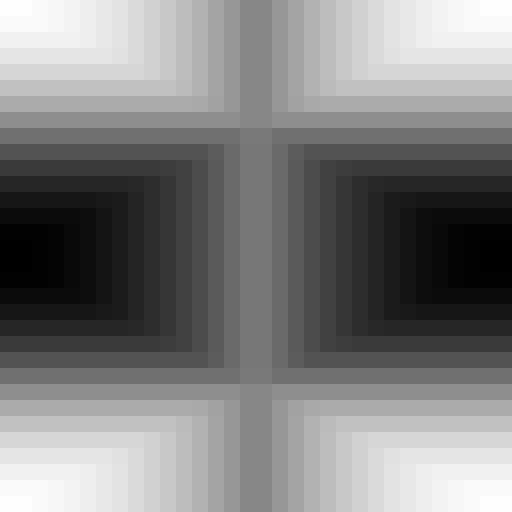} \includegraphics[height=20mm]{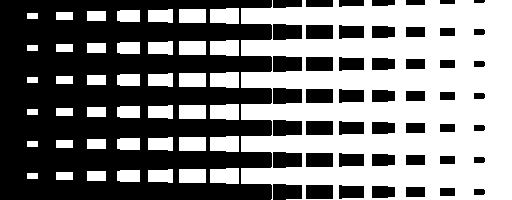}\\
(b) \includegraphics[height=20mm]{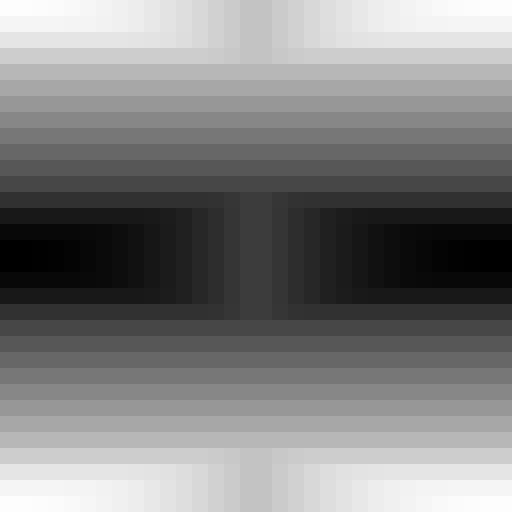} \includegraphics[height=20mm]{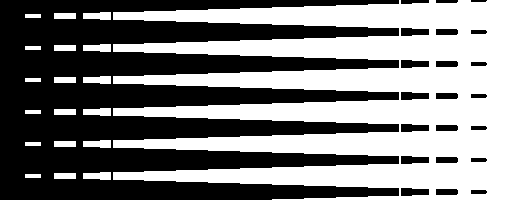}\\
(c) \includegraphics[height=20mm]{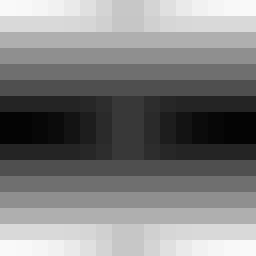} \includegraphics[height=20mm]{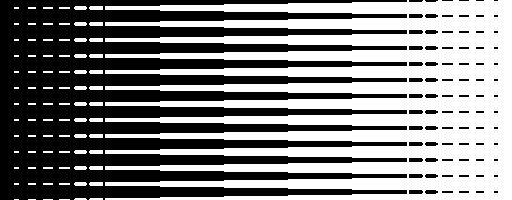}\\
(d) \includegraphics[height=20mm]{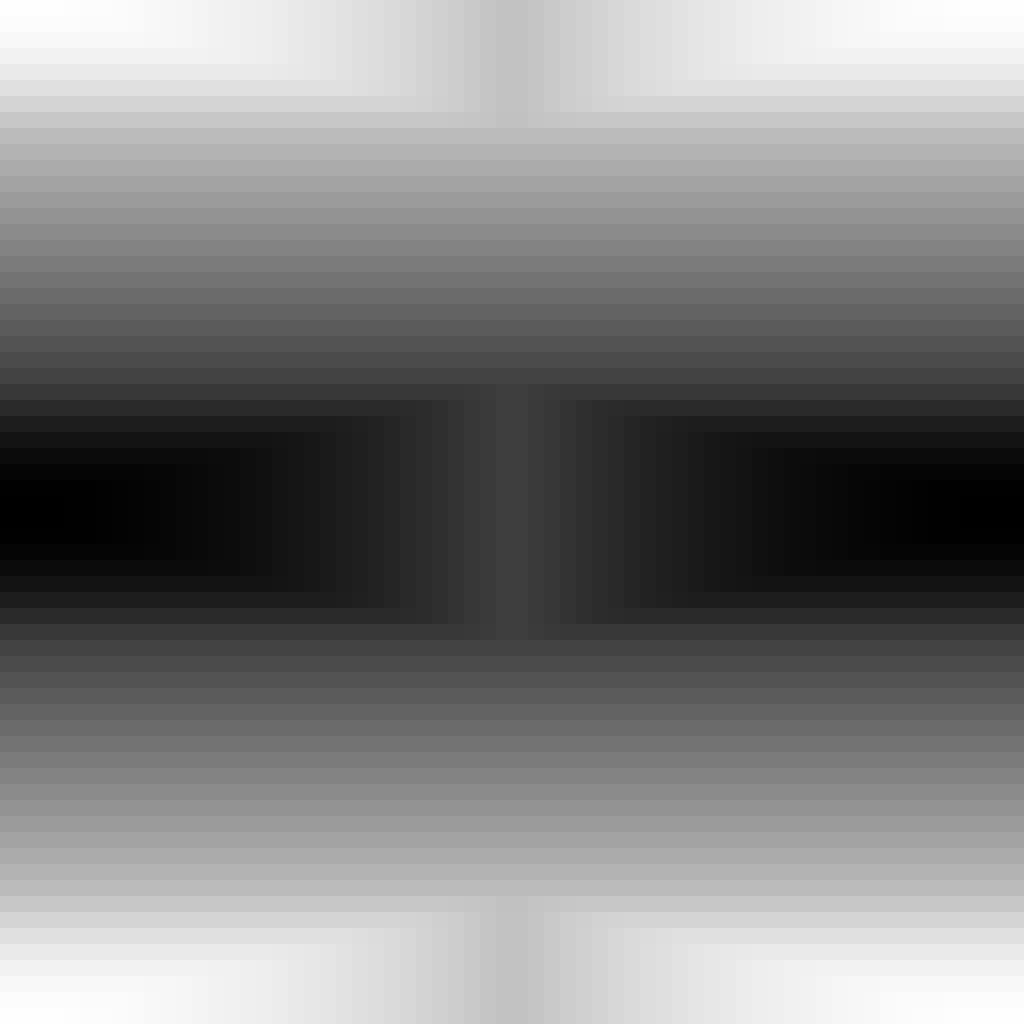} \includegraphics[height=20mm]{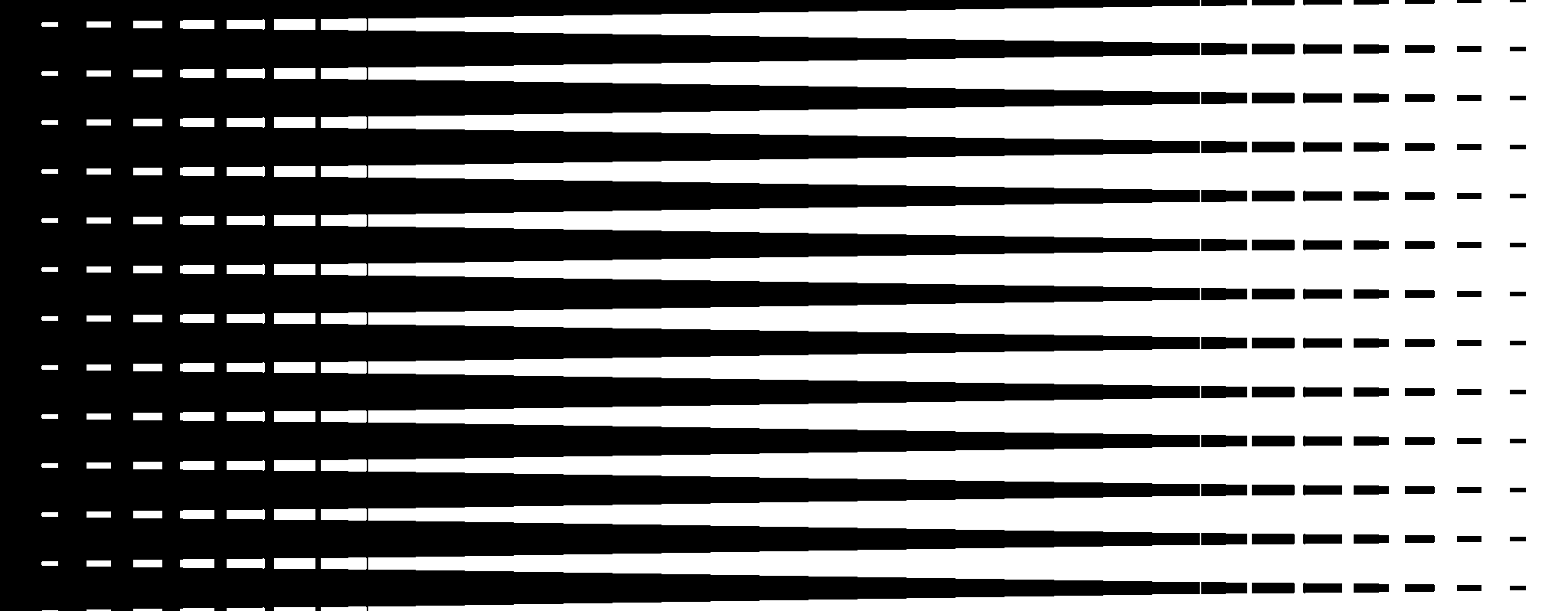}\\
(e) \includegraphics[height=20mm]{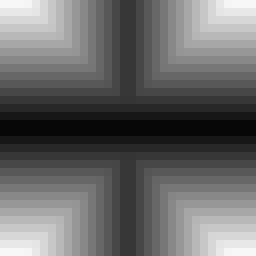} \includegraphics[height=20mm]{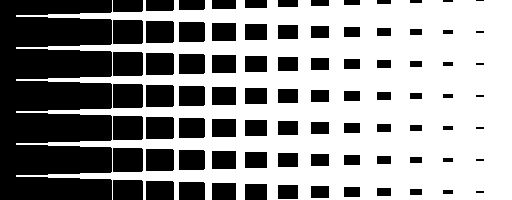}
\caption{Application of the cross-hatching dither matrix to an intensity ramp:
a) $S=0.5$;
b) $S=0.25$;
c) $S=0.25$ and the dither matrix is halved in size;
d) $S=0.25$ -- compared to c) both the dither matrix and intensity ramp are upsampled by a factor of 4, and the dithered result is downsampled by 4;
e) dither matrix from Ostromoukhov~\cite{Ostromoukhov}.}
\label{dither2}
\end{figure}

Figure~\ref{dither2}a and~b shows the effect of using values of $S=0.5$ and $S=0.25$ for constructing the dither matrices.
We prefer to use mainly horizontal lines with only a small
amount of cross hatching in order to keep the results simpler and less cluttered, and so we use $S=0.25$.
Figure~\ref{dither2}c also shows how reducing the dimensions of the dither matrix (simple averaging is used)
increases the frequency of the engraving lines.
However, when the dither matrix becomes low resolution the digitization artifacts become visible in the dithered result;
each horizontal line appears as a concatenation of rectangles rather than an elongated triangle.
This can be avoided by upsampling the intensity image, applying a similarly upsampled dither matrix
(i.e. less downsampled from the original high resolution dither matrix),
and then downsampling the dithered result using interpolation.
As shown in figure~\ref{dither2}d, the result, which is now grayscale rather than binary, appears more continuous.

For comparison, figure~\ref{dither2}e shows the dither matrix ``L1 smaller (L2 raised 3/16)'' from Ostromoukhov~\cite{Ostromoukhov}.
It has some of the characteristics that we are aiming for,
i.e. lines rather than the spots which are produced by his other merging mode examples,
and also it generates cross-hatching lines.
However, there are only white cross-hatching lines;
the black cross-hatching lines are missing.

\section{Portrait Engraving}

The dither matrix constructed in section~\ref{dithering} allows an image to be re-rendered as
an engraving with horizontal black and white lines with vertical cross-hatchings.
Now we describe an additional step which warps these lines so as to enhance the form of the face.
If the 3D geometry of the scene was available, then this could be used to control the orientation of the engraving lines.
Since the 3D information is not directly known, it can instead be estimated.
However, if there are errors in the estimated 3D geometry this could lead to a mismatch between the orientation and thickness
cues present in the engraving lines.
To avoid an ``uncanny valley'' effect, instead we use the simple and rough proxy geometry
that was previously described in Rosin and Lai's portrait stylisation algorithm~\cite{rosin-portrait} for creating a shading effect.

\begin{figure}[htbp]
\centering
\subfigure[]{\includegraphics[height=30mm]{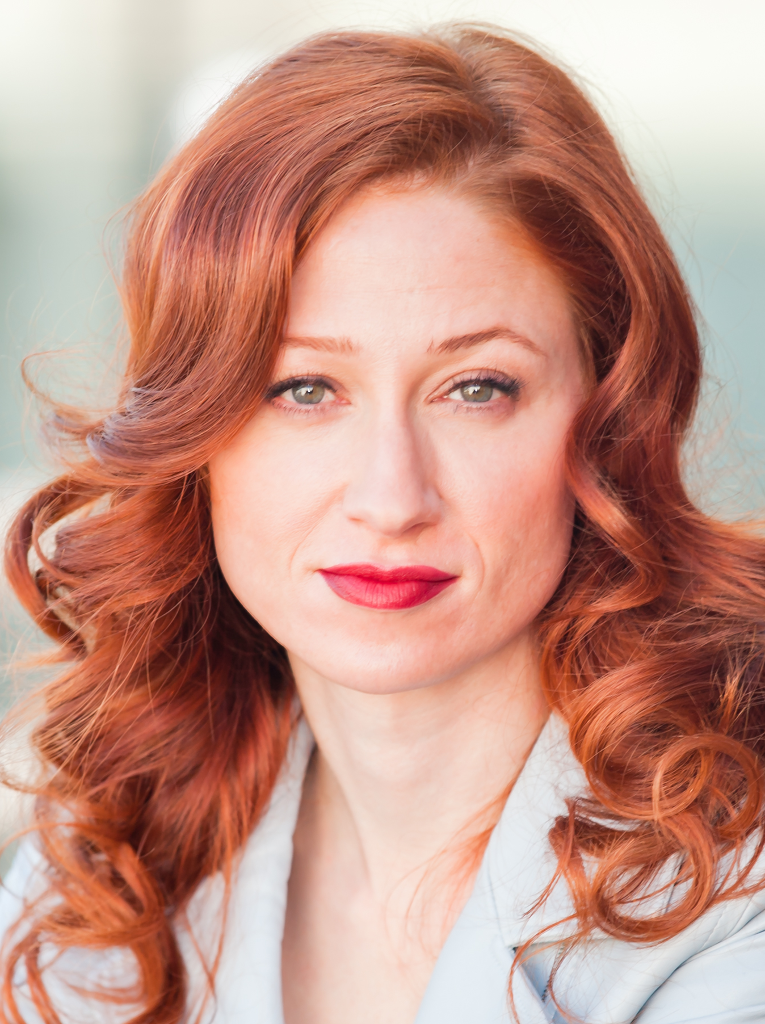}}
\subfigure[]{\includegraphics[height=30mm]{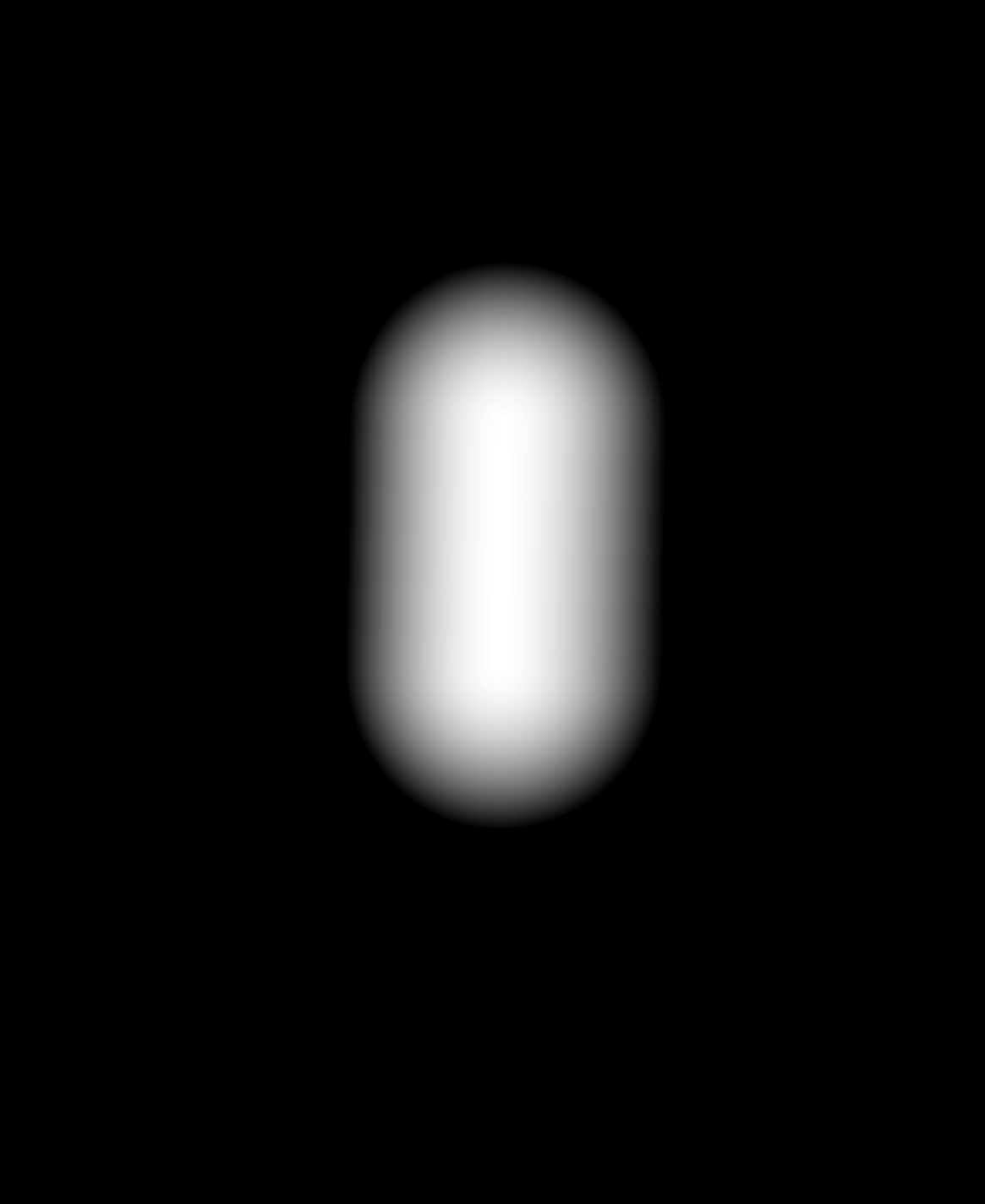}}
\subfigure[]{\includegraphics[height=30mm]{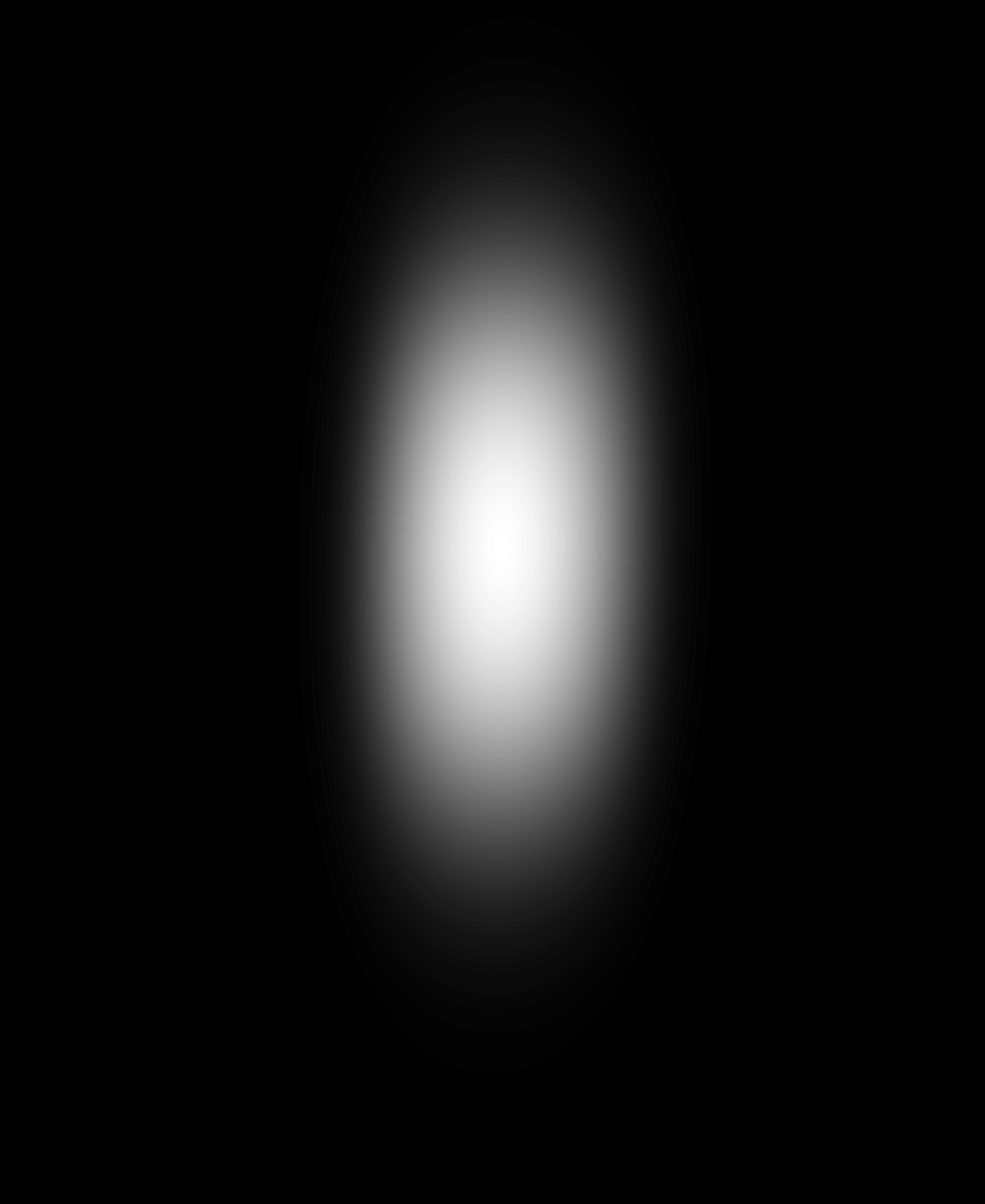}}

\subfigure[]{\includegraphics[height=30mm]{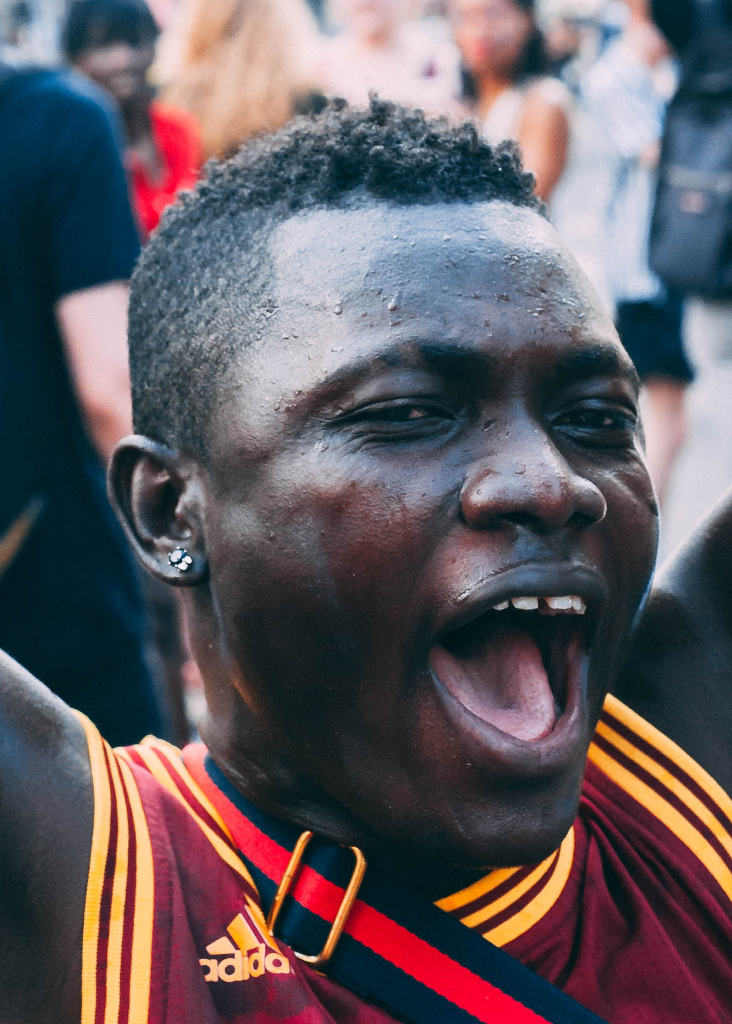}}
\subfigure[]{\includegraphics[height=30mm]{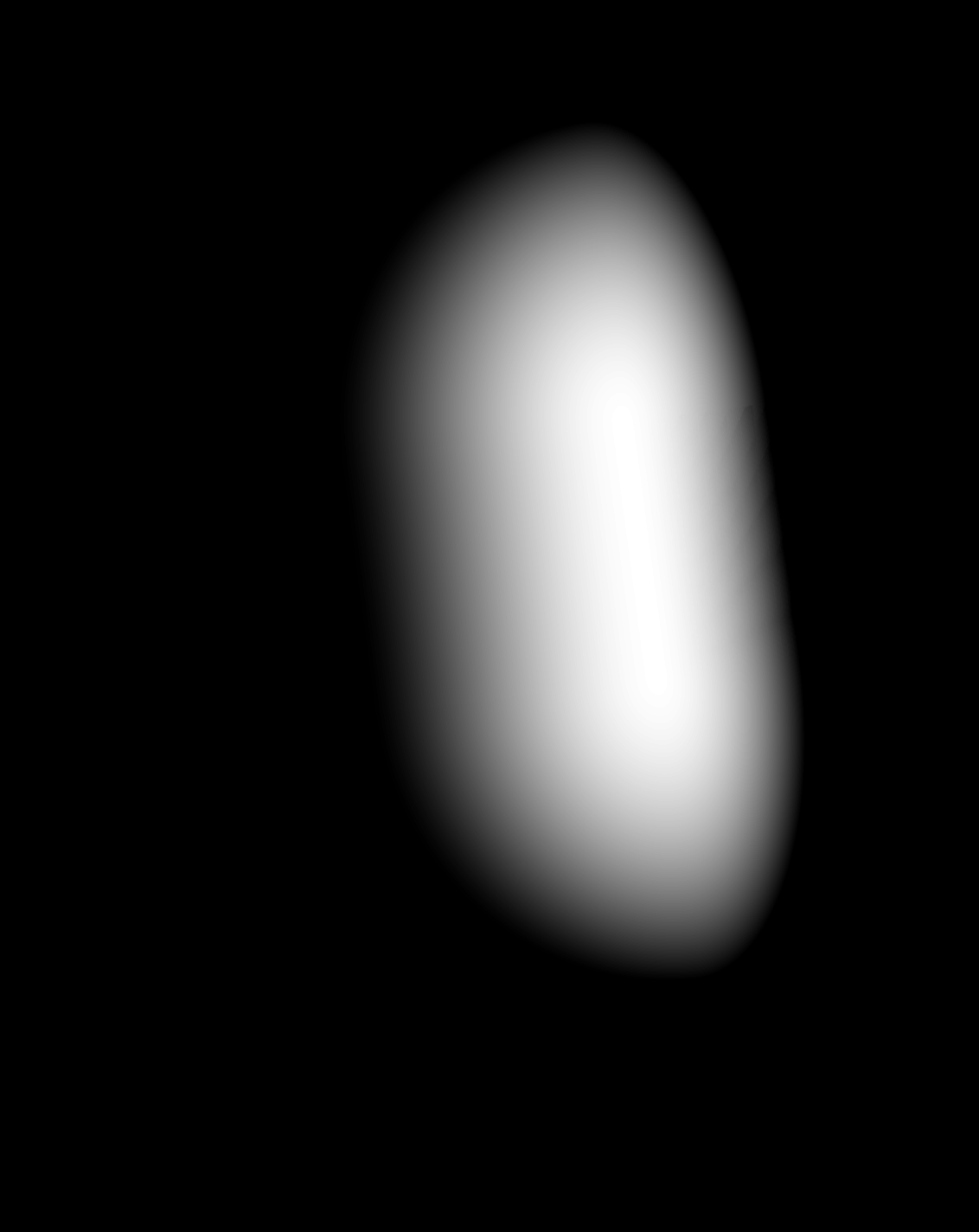}}
\subfigure[]{\includegraphics[height=30mm]{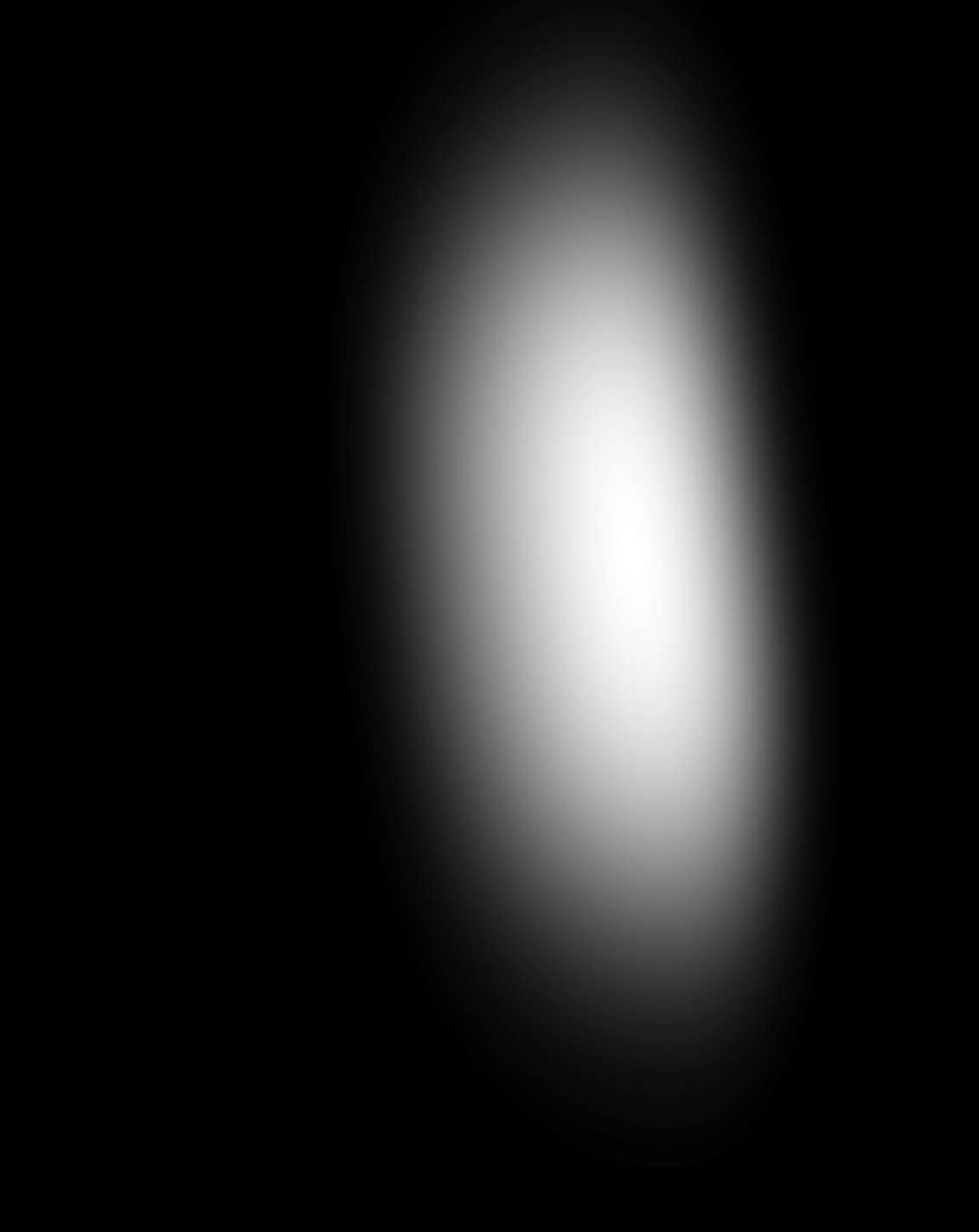}}

\subfigure[]{\includegraphics[height=70mm]{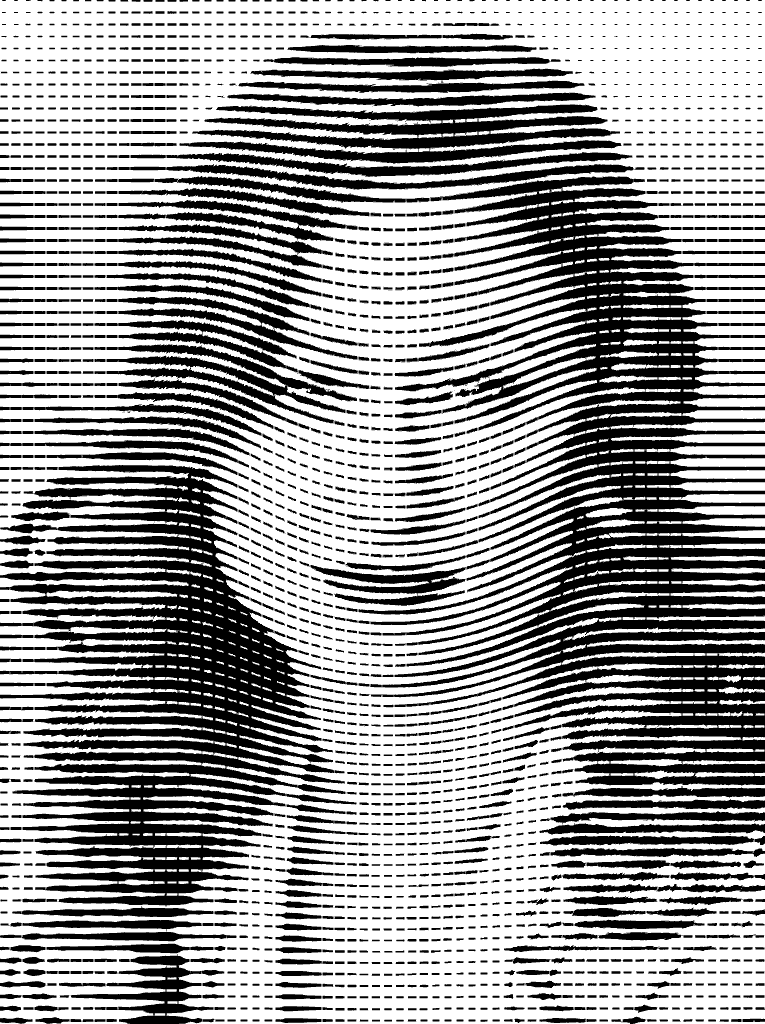}}
\subfigure[]{\includegraphics[height=70mm]{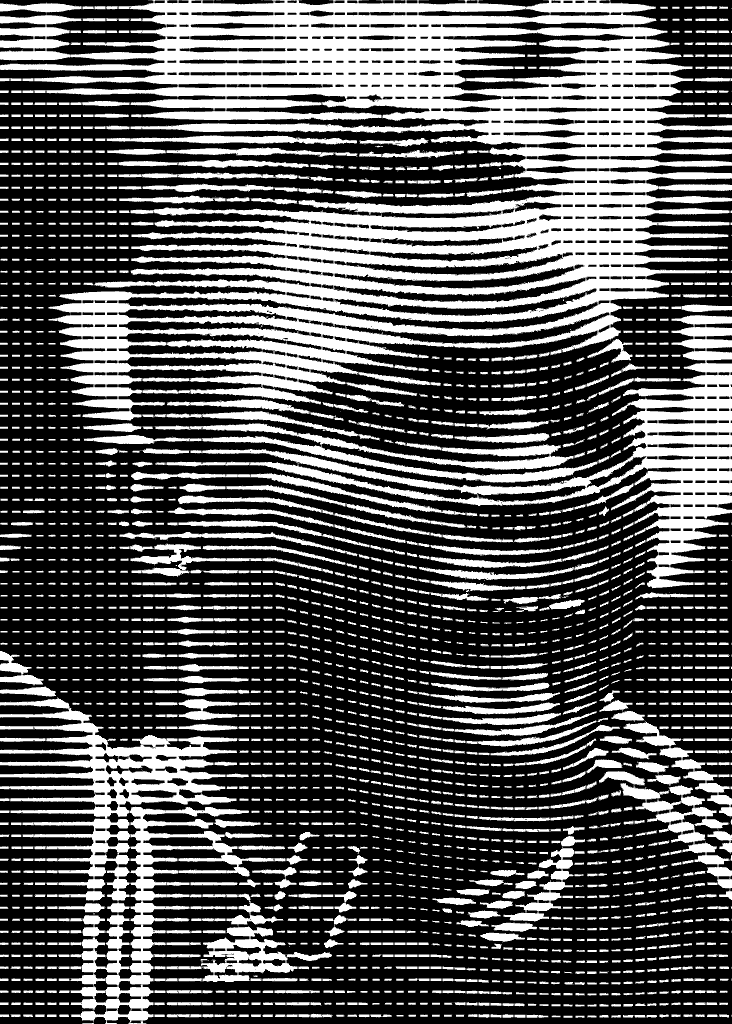}}
\caption{Generated Shading images before (b\&e) and after (e\&f) anisotropic blurring;
(g\&h) are the final engravings.
}
\label{shading}
\end{figure}

\section{Portrait Engraving}

First, a facial model is fitted to the image using OpenFace~\cite{baltru2016openface},
and this is used to extract a mask outlining the extent of the face.
Since the model is missing the upper part of the face (i.e. above the eyebrows),
it is extended upwards by a fixed proportion and closed to provide an approximate mask.
Shading is applied using a simplified Phong reflection model, where we focus only on the ambient and diffusion terms,
with the pseudo-specular effect added later.
We further simplify the problem by assuming a frontal lighting direction.
Rather than use a true three dimensional model for the face surface we approximate the angle $\theta$
that a normal makes with the frontal lighting direction by computing
a distance field from an extended version of the bridge of the nose, which is obtained from the fitted face model.
The distances are rescaled so that they reach the value $\frac{\pi}{2}$ at the border of the face mask
(i.e. these have normal directions orthogonal to the lighting direction).
For points in the face mask this produces a reasonable simple surface that is approximately a cylinder with rounded ends.
The pseudo-normals for all points outside of the face mask are truncated to $\frac{\pi}{2}$.
Shading is then generated as $\left(\cos \theta (1-\alpha) + \alpha \right)$,
where $\alpha=0.4$ is the weight for the ambient light, and $1-\alpha$ is the weight for the diffusion term.
To avoid discontinuities in the engraving lines at the face boundary,
as well as artifacts at the top of the head due to the simple synthesis of the upper face,
the shading image is blurred, especially at the top and bottom.
This is done by applying anisotropic Gaussian filtering in the direction of the head orientation estimated by OpenFace~\cite{geusebroek2003fast}.
Figure~\ref{shading} demonstrates these steps.

\begin{figure}[htbp]
\centering
\subfigure[]{\includegraphics[height=70mm]{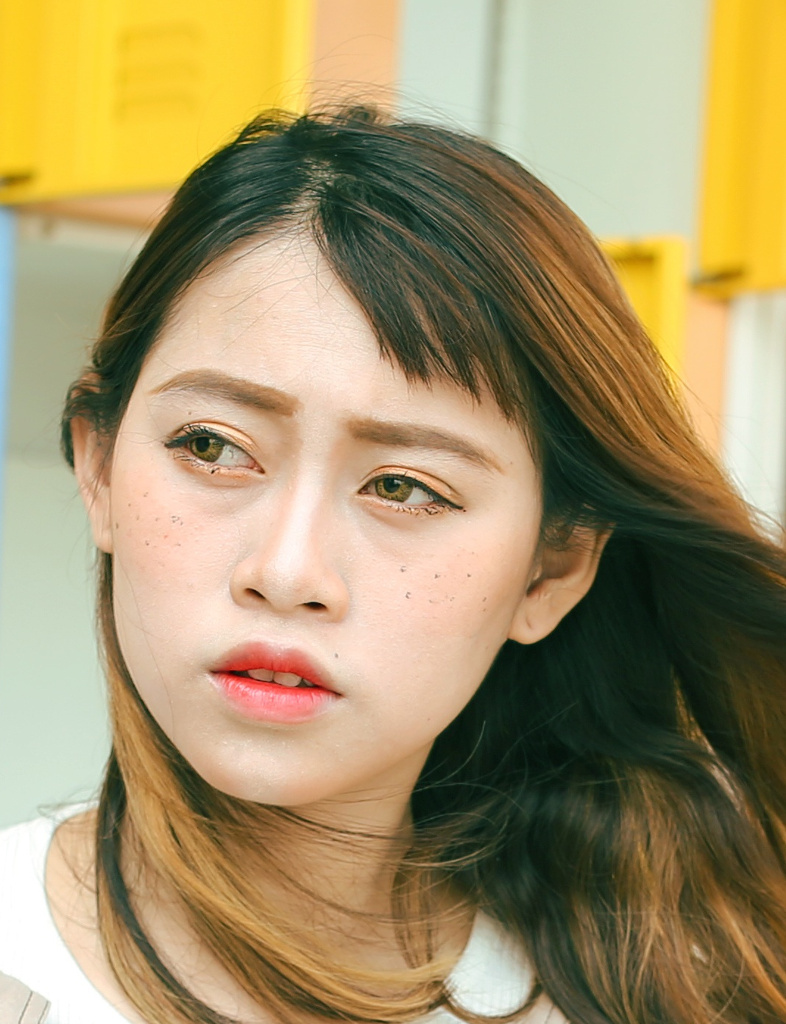}}
\subfigure[]{\includegraphics[height=70mm]{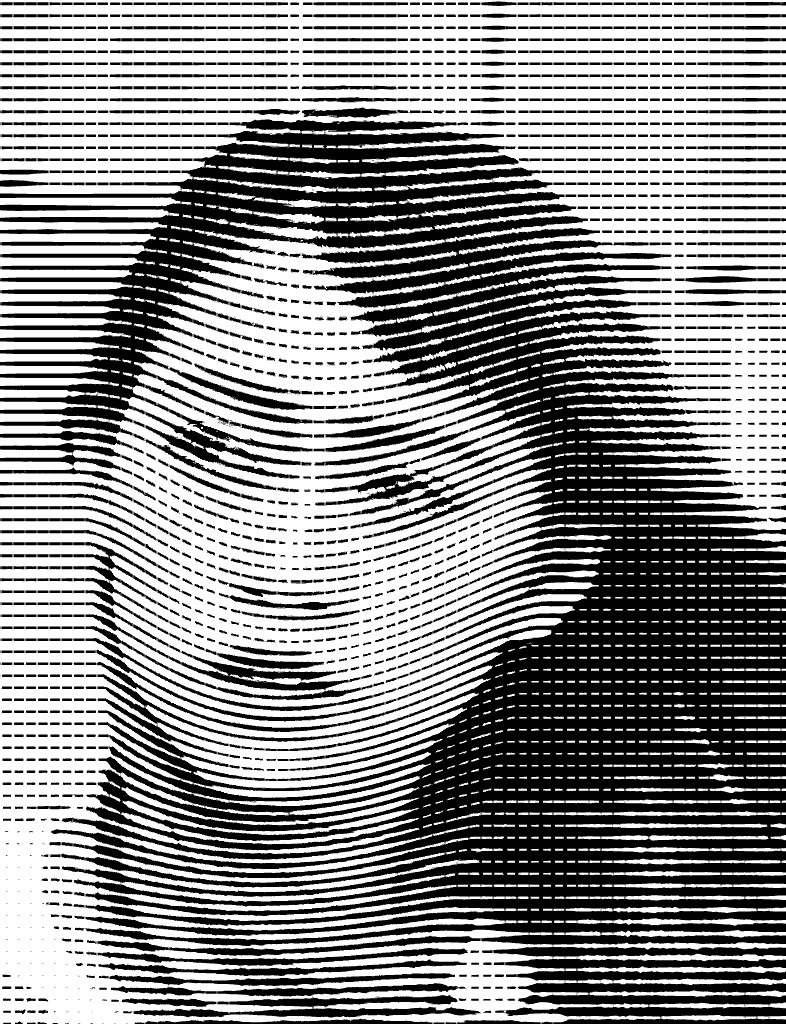}}
\subfigure[]{\includegraphics[height=70mm]{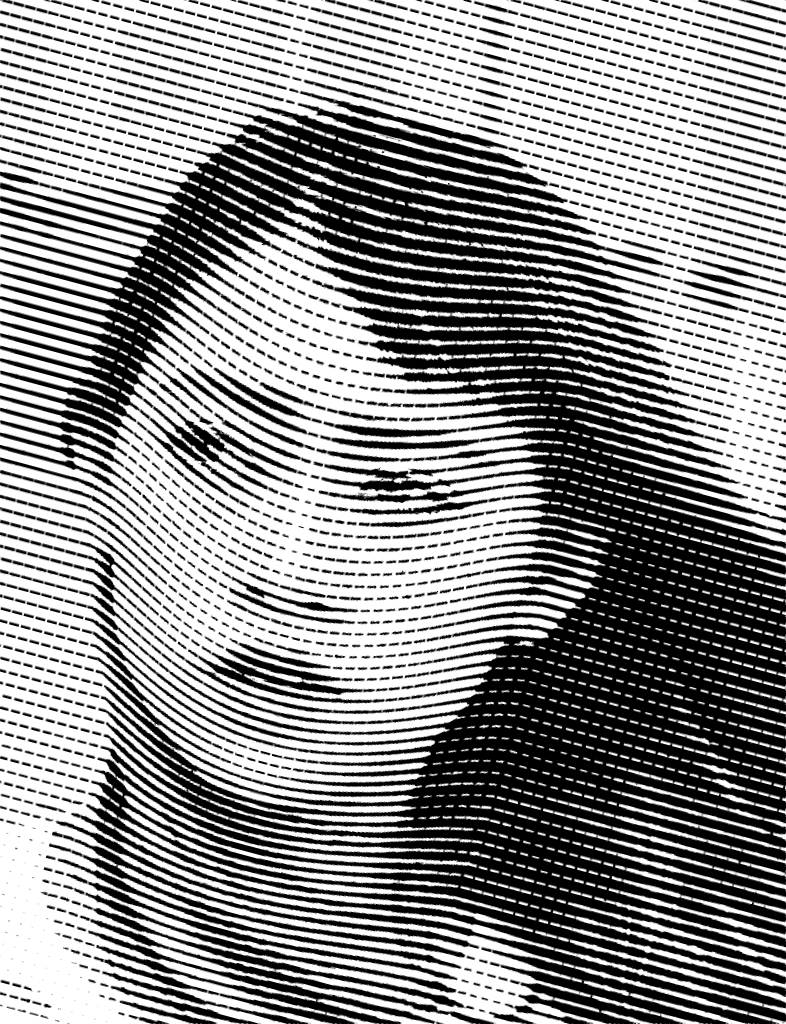}}
\caption{The engraving stylisation has been applied to the input image in (a)
using a purely vertical warp to generate the curved engraving lines in (b).
In (c) the curved engraving lines are aligned with the face by incorporating
pre- and post-rotation to the dithering.
}
\label{rotate}
\end{figure}

Whereas in Rosin and Lai's algorithm~\cite{rosin-portrait} the shading image was applied to the stylised image to create a shading effect,
for engraving we use it instead to warp the dither matrix so that the lines curve around the face, providing a pseudo-3D effect.
For its current application the portrait engraving has been applied to faces that are roughly frontal facing and vertically aligned.
Therefore, a simple one dimensional warp is sufficient:
each pixel in the shading image defines a vertical offset that is applied to warp a version of the
dither matrix that has been tiled to cover the portrait image.
Ordered dithering is then applied as before.
If it is desirable that the warp is aligned with the face orientation, then a two-dimensional warp vector could be defined instead;
alternatively, the image could be rotated to vertically align the face before dithering, and the resulting engraving rotated back, as shown in figure~\ref{rotate}.

\begin{figure}[htbp]
\centering
\subfigure[]{\includegraphics[height=35mm]{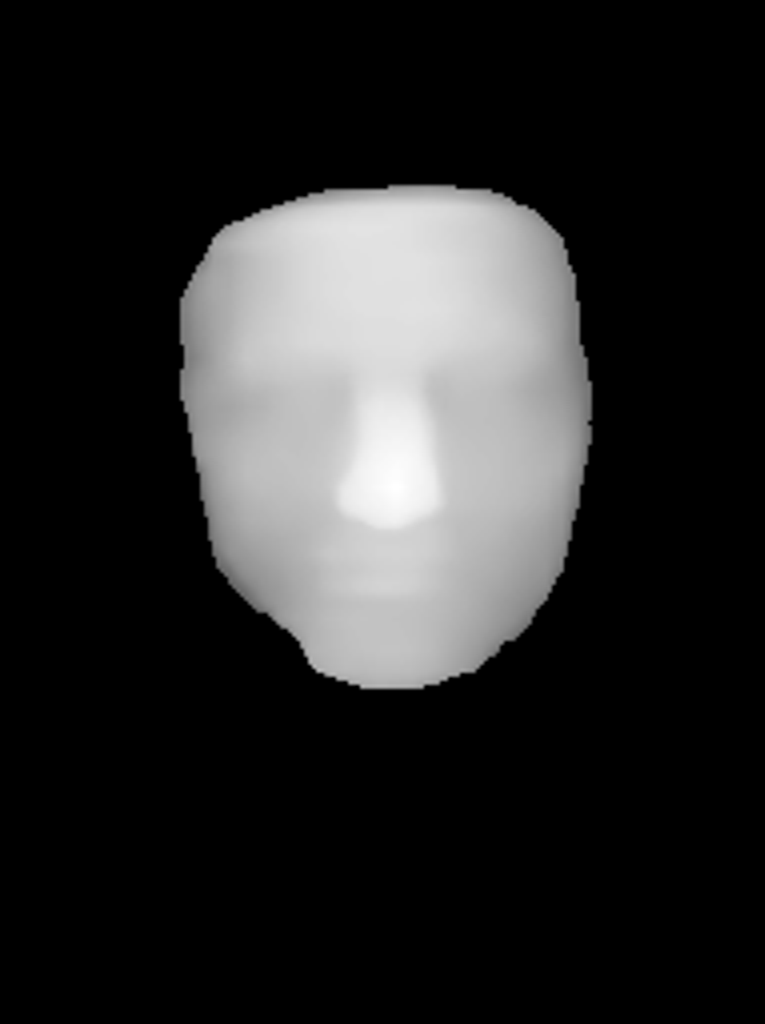}}
\subfigure[]{\includegraphics[height=70mm]{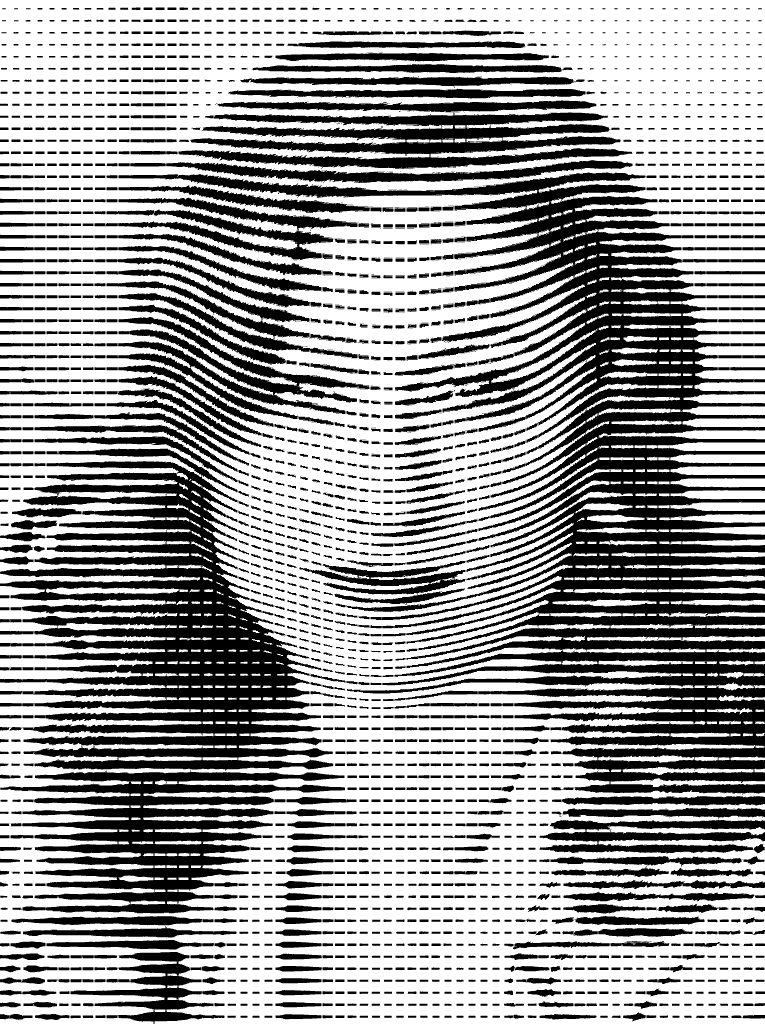}}
\subfigure[]{\includegraphics[height=35mm]{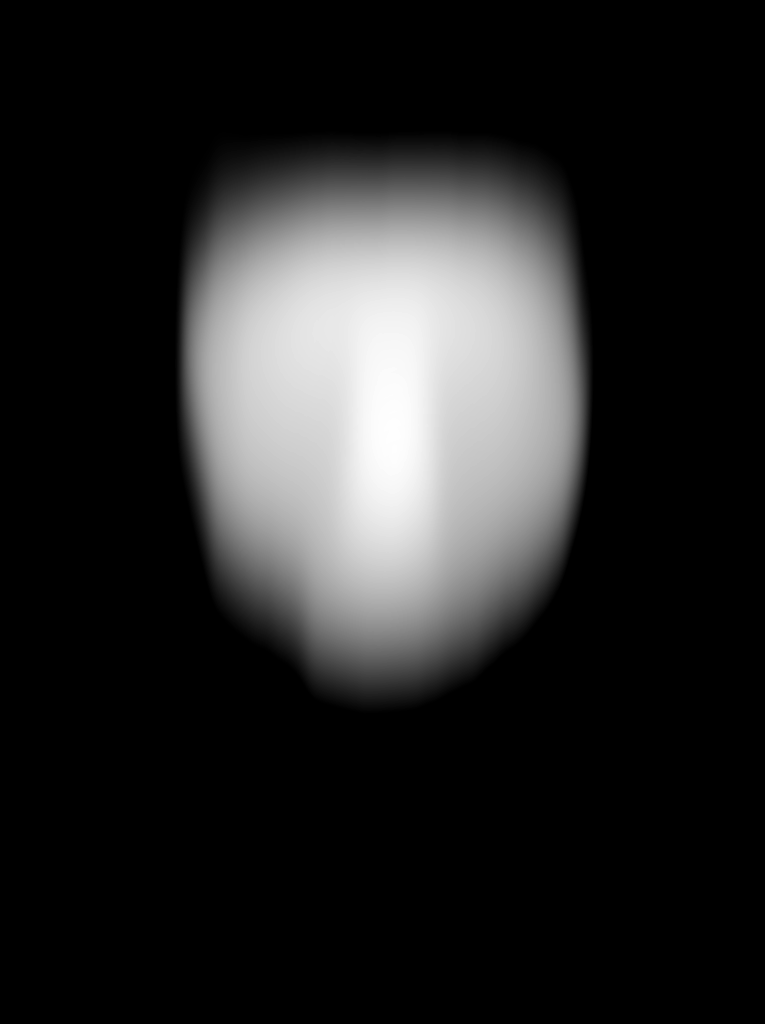}}
\subfigure[]{\includegraphics[height=70mm]{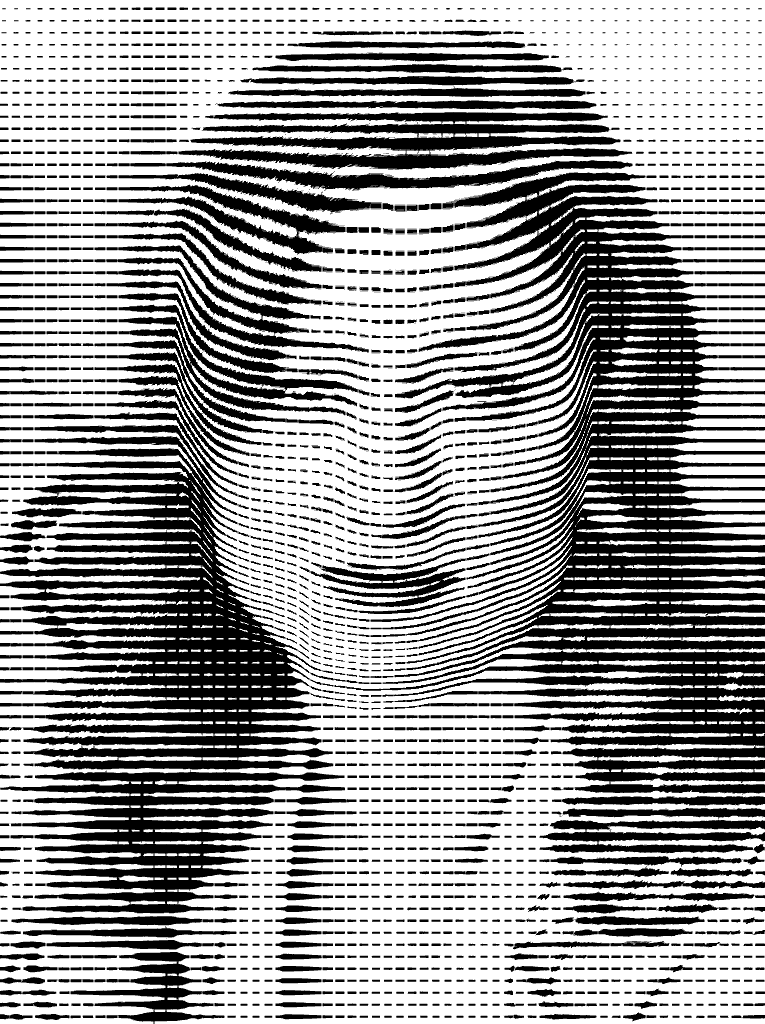}}
\caption{Engraving using an estimated depth image instead of the shading image.
}
\label{depth}
\end{figure}

An alternative to using a cylinder as the rough proxy geometry for the head is to replace it with an existing depth estimation method.
We have applied the 3D face reconstruction approach in~\cite{jackson2017large} which uses a
CNN to perform regression of a volumetric representation from a single 2D image.
From this, a depth map was extracted, shown in figure~\ref{depth}a.
The existing pipeline, including the substantial anisotropic blurring, was run, and the result
in figure~\ref{depth}b is qualitatively very similar to that in figure~\ref{shading}.
Less blurring has been applied to the depth map in figure~\ref{depth}c in an attempt to retain more detail,
e.g. the bridge of the nose is preserved.
However, the engraving result (figure~\ref{depth}d) is less satisfactory;
although the curvature of the nose is more evident, the sides of the face which have a high gradient,
are not well integrated with the remainder of the engraving.

\begin{figure}[htbp]
\subfigure[]{\includegraphics[height=70mm]{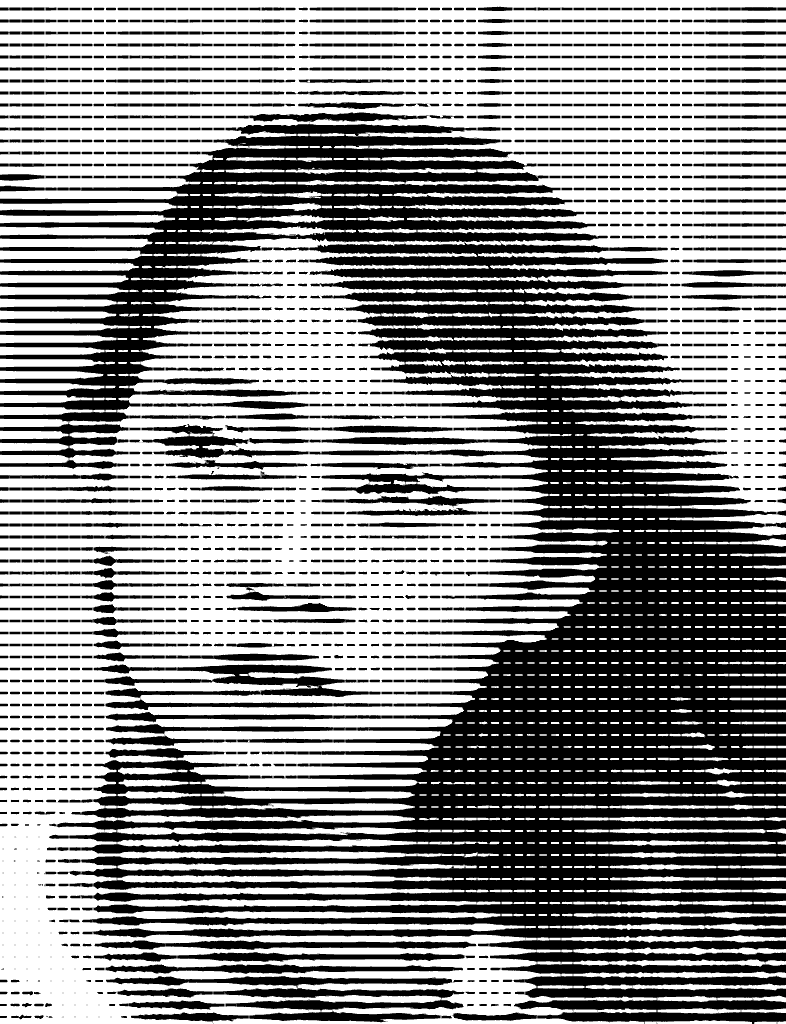}}
\subfigure[]{\includegraphics[height=70mm]{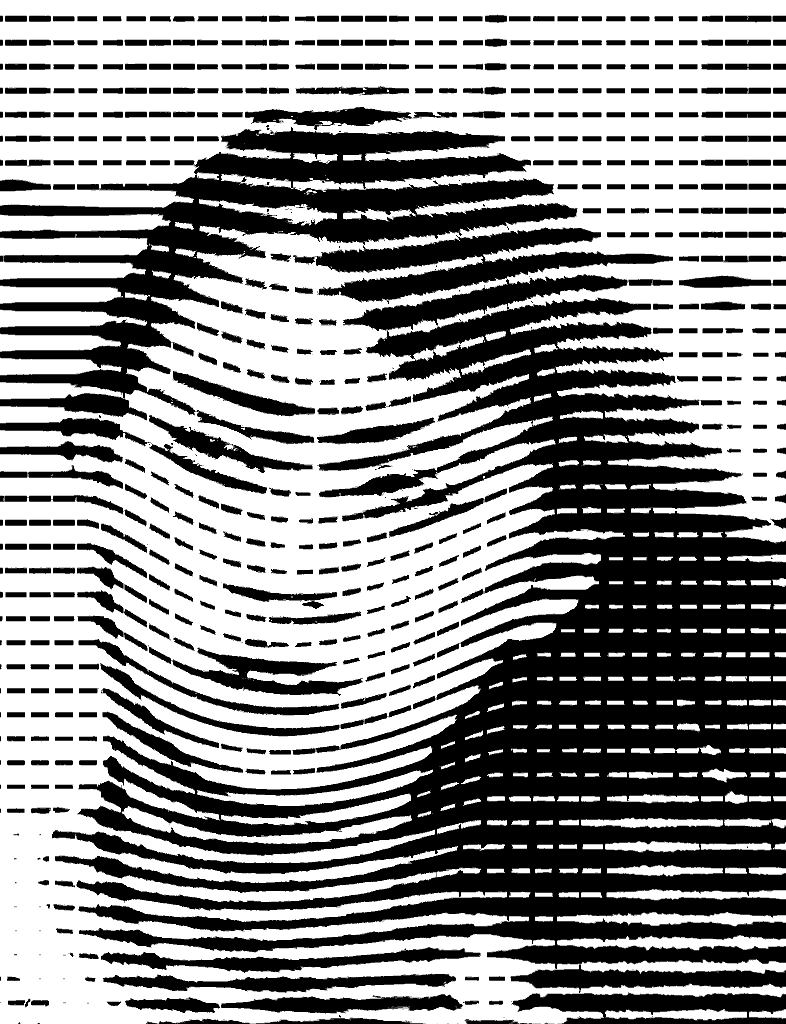}}
\subfigure[]{\includegraphics[height=70mm]{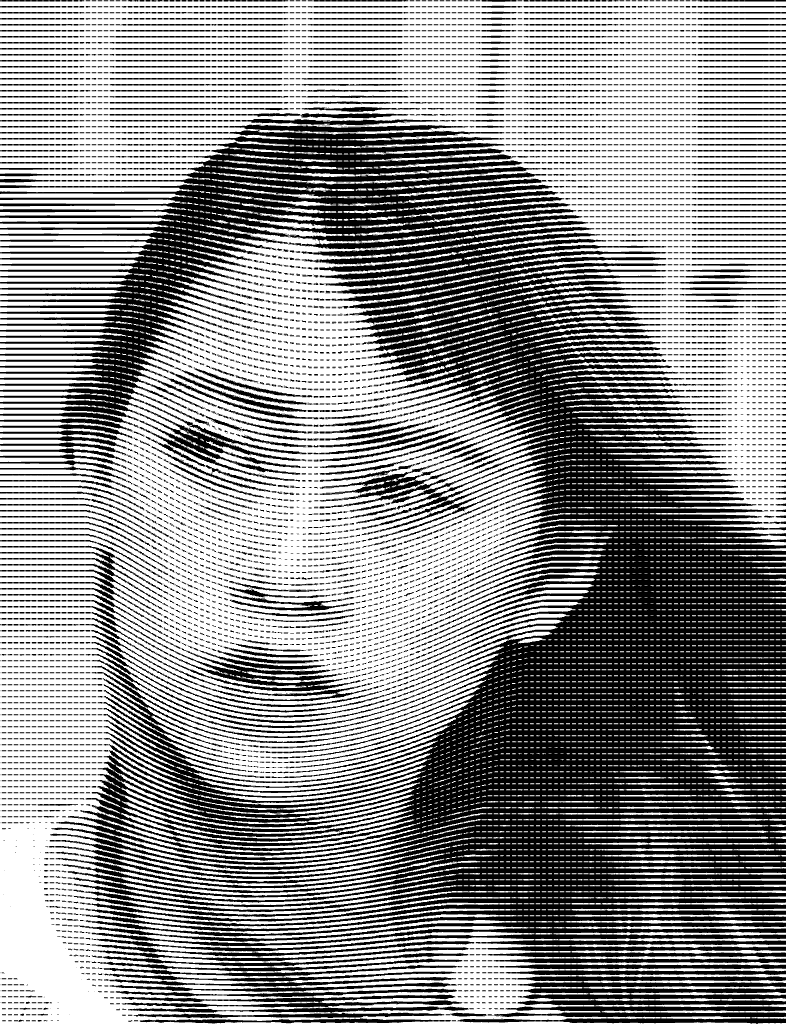}}
\caption{Variations in engraving effects.
a) using straight rather than curved engraving lines,
b) lower density of engraving lines,
c) higher density of engraving lines.
}
\label{variants}
\end{figure}

Figure~\ref{variants} shows alternative engraving stylisations of the image in figure~\ref{rotate}a.
In figure~\ref{variants}a no warping has been applied to the dither matrix, and so all the engraving lines are straight.
It can be seen that there is a loss of expressiveness compared to figure~\ref{rotate}b.
The dither matrix has been upsampled and downsampled by a factor of two, produced the results in figure~\ref{variants}b\&c
in which the engraving line density is altered.


\section{Colour Engraving}

\begin{figure}[htbp]
\centering
\subfigure[]{\includegraphics[width=0.28 \textwidth]{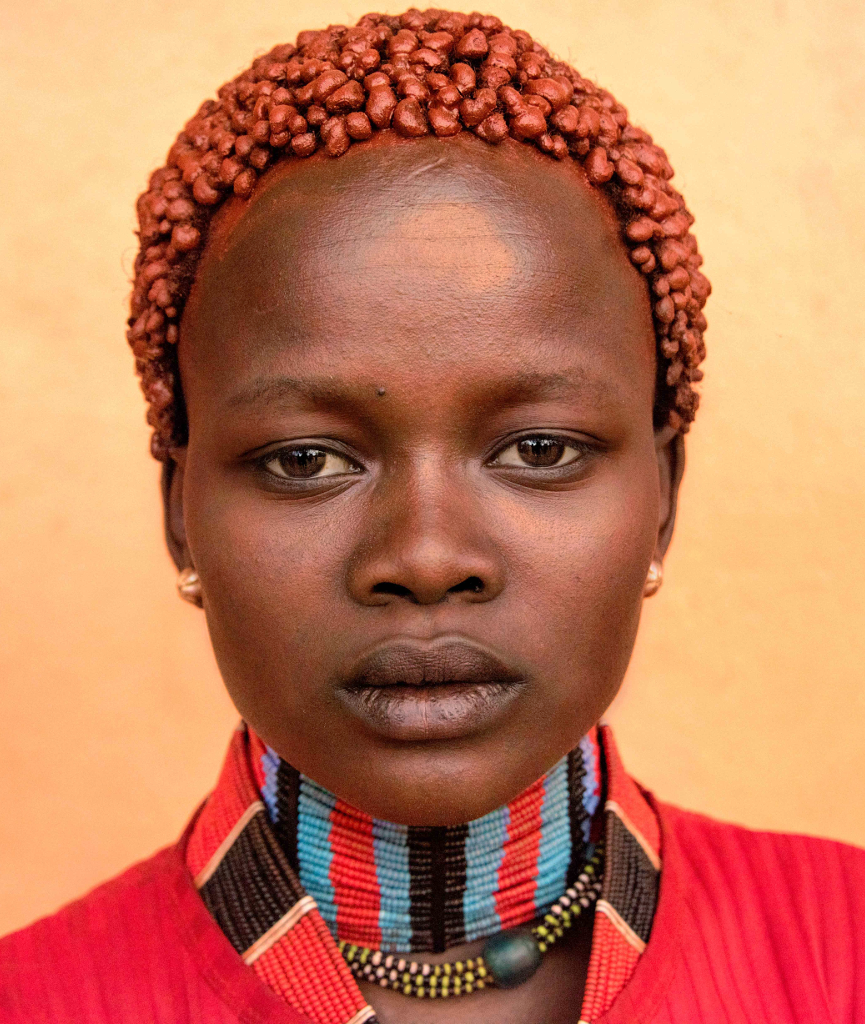}}
\subfigure[]{\includegraphics[width=0.28 \textwidth]{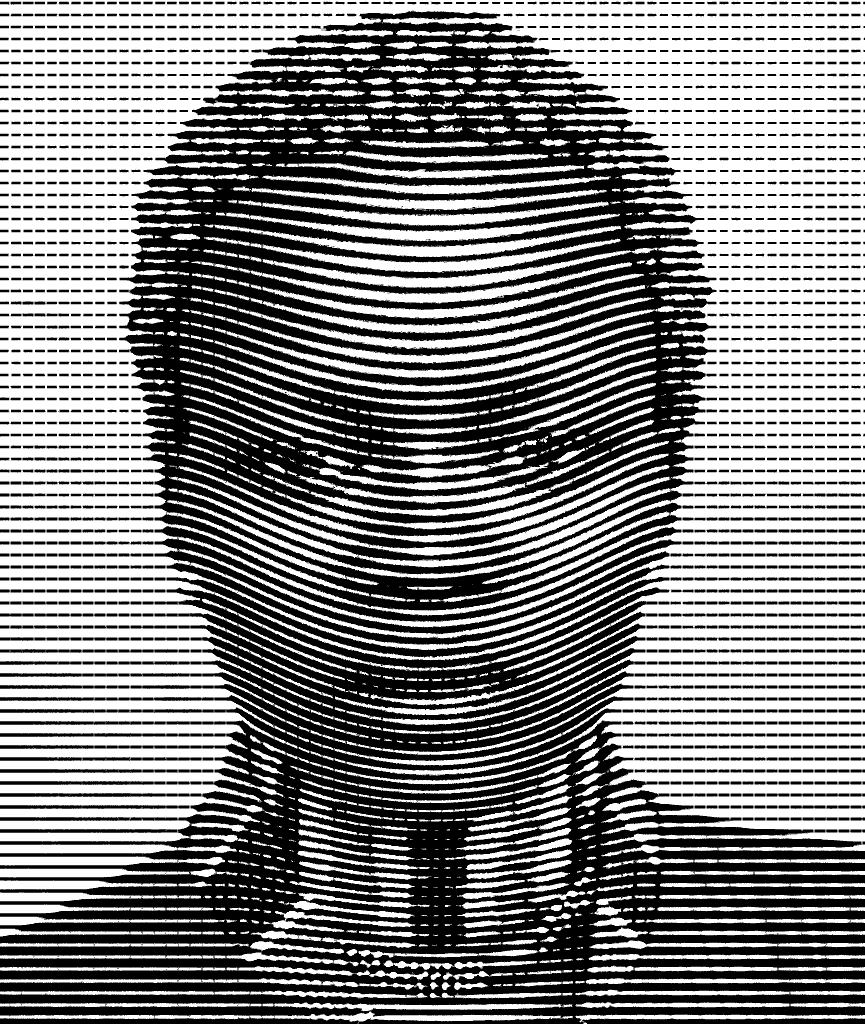}}
\subfigure[]{\includegraphics[width=0.28 \textwidth]{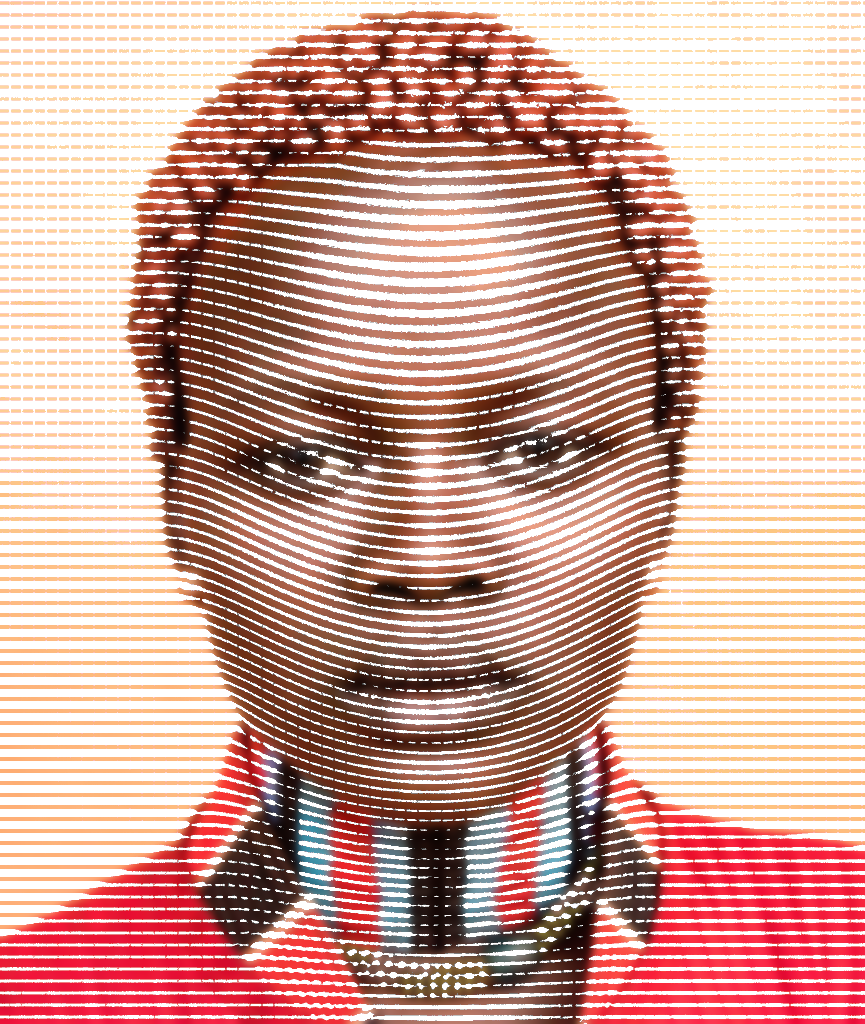}}
\subfigure[]{\includegraphics[width=0.28 \textwidth]{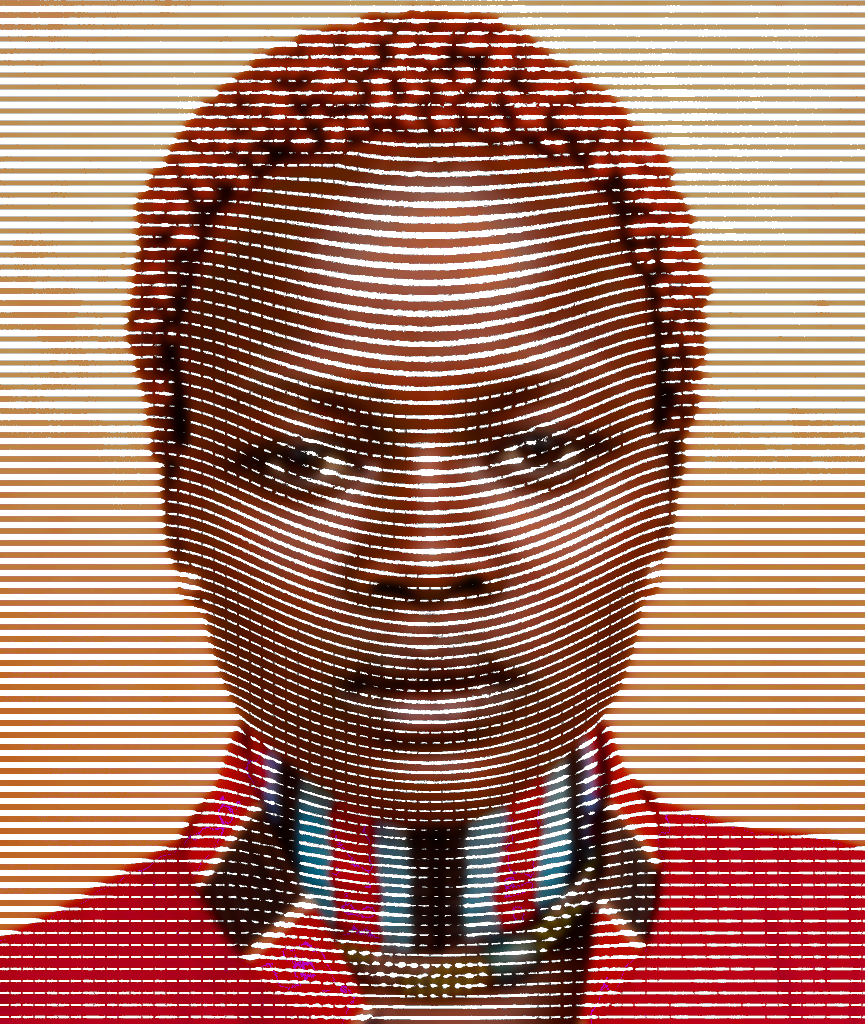}}
\subfigure[]{\includegraphics[width=0.28 \textwidth]{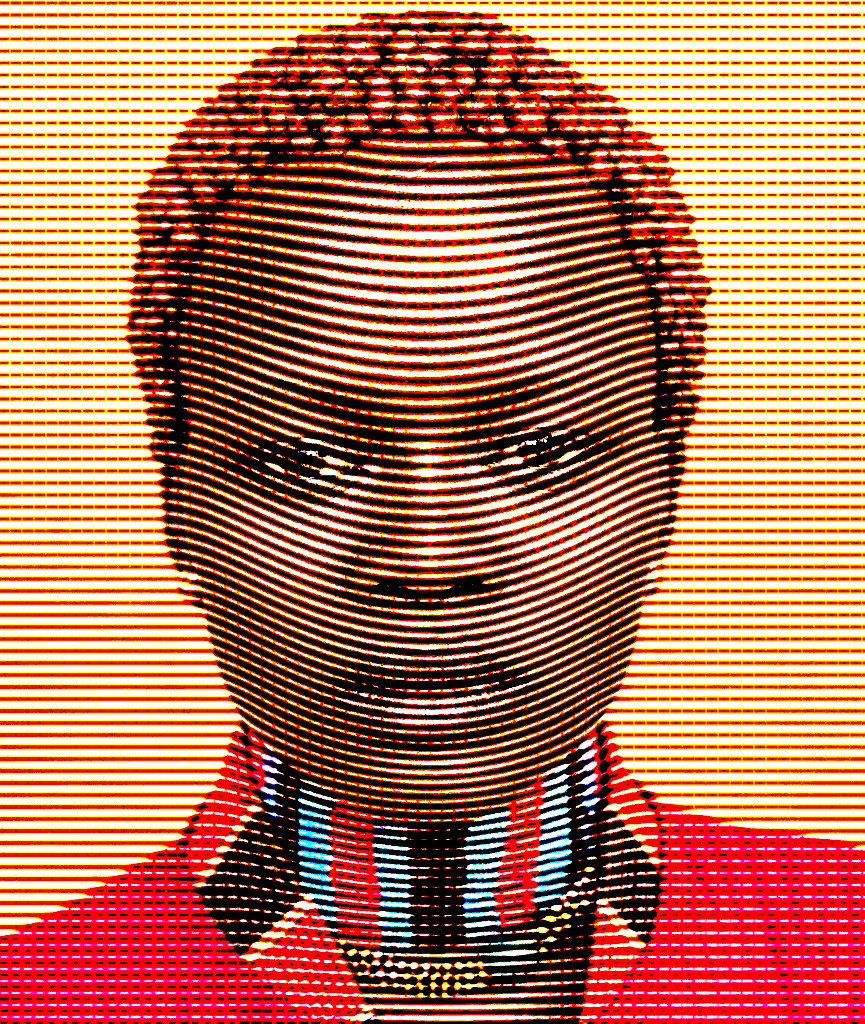}}
\subfigure[]{\includegraphics[width=0.28 \textwidth]{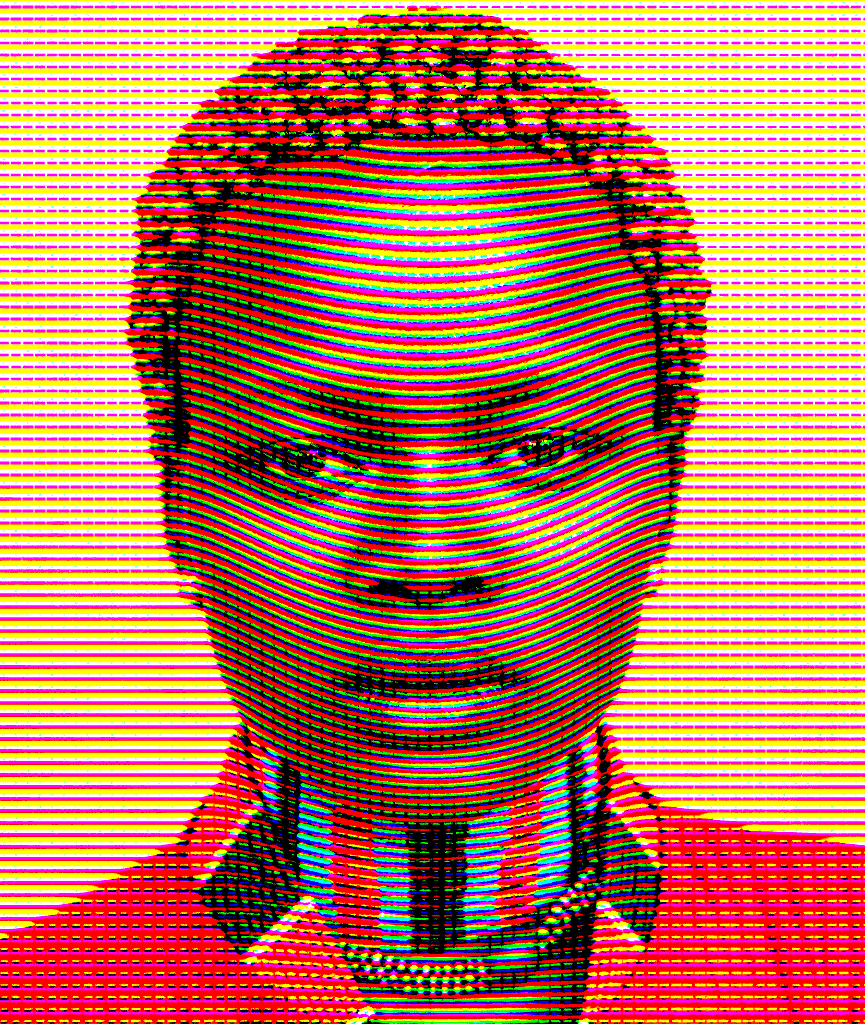}}
\subfigure[]{\includegraphics[width=0.28 \textwidth]{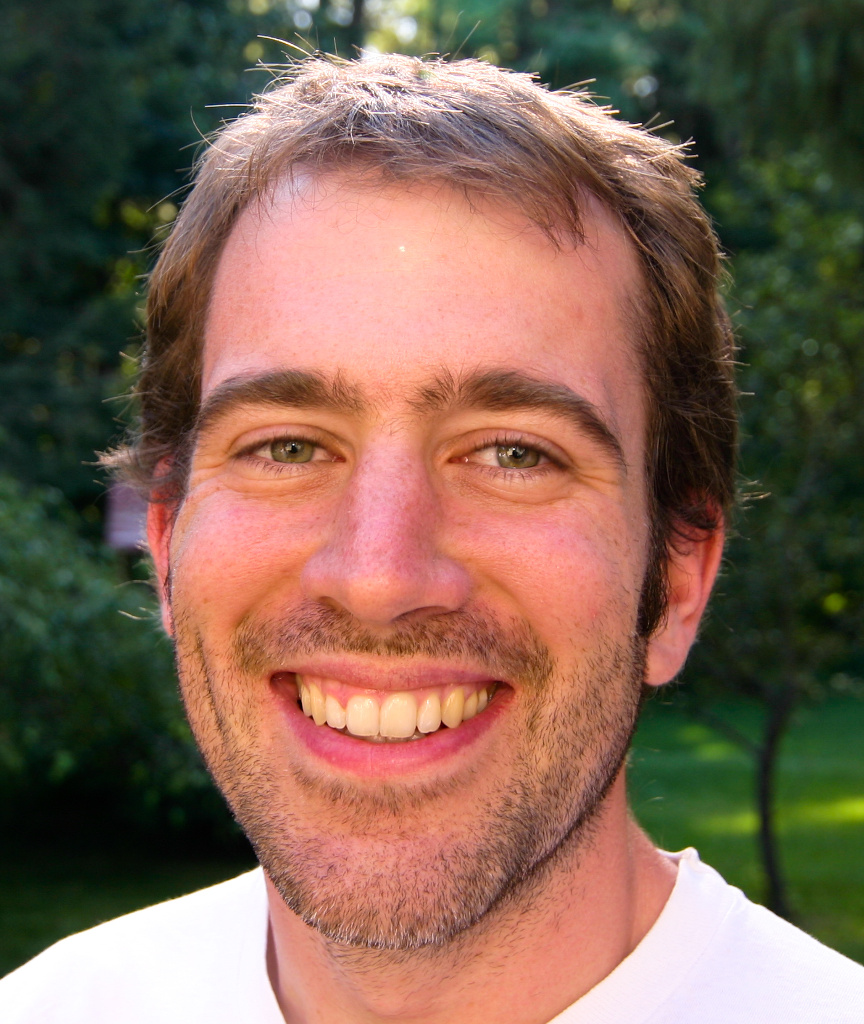}}
\subfigure[]{\includegraphics[width=0.28 \textwidth]{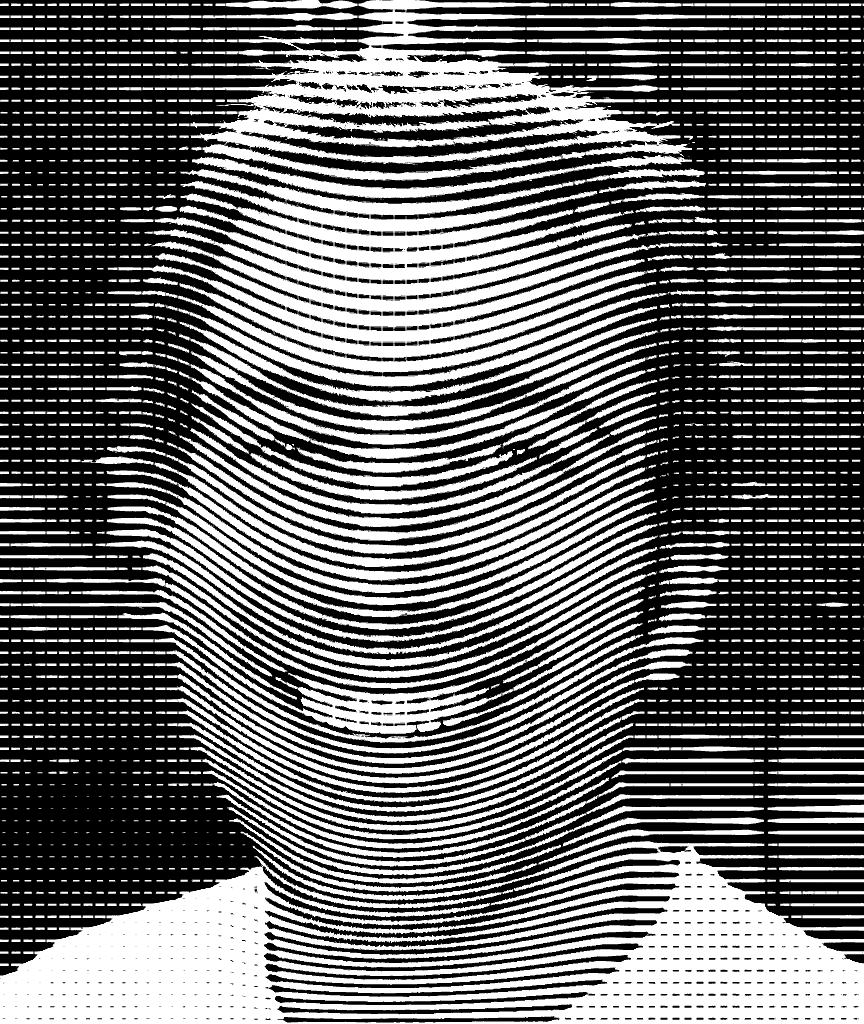}}
\subfigure[]{\includegraphics[width=0.28 \textwidth]{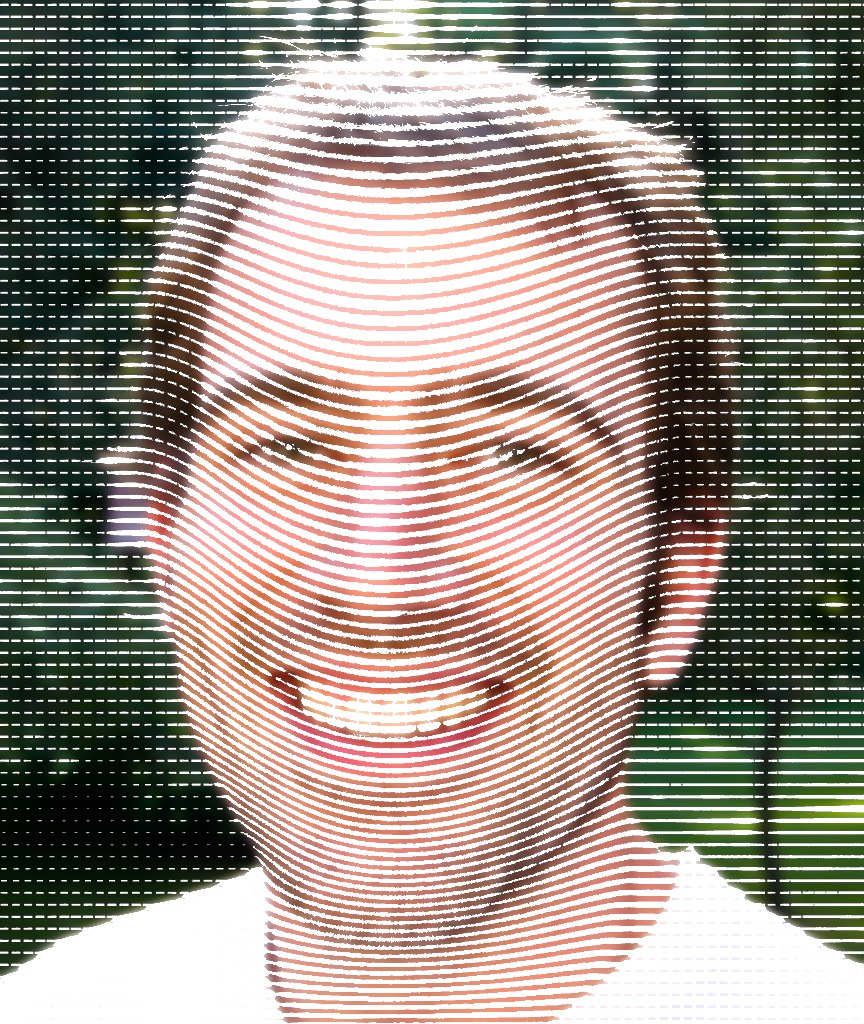}}
\subfigure[]{\includegraphics[width=0.28 \textwidth]{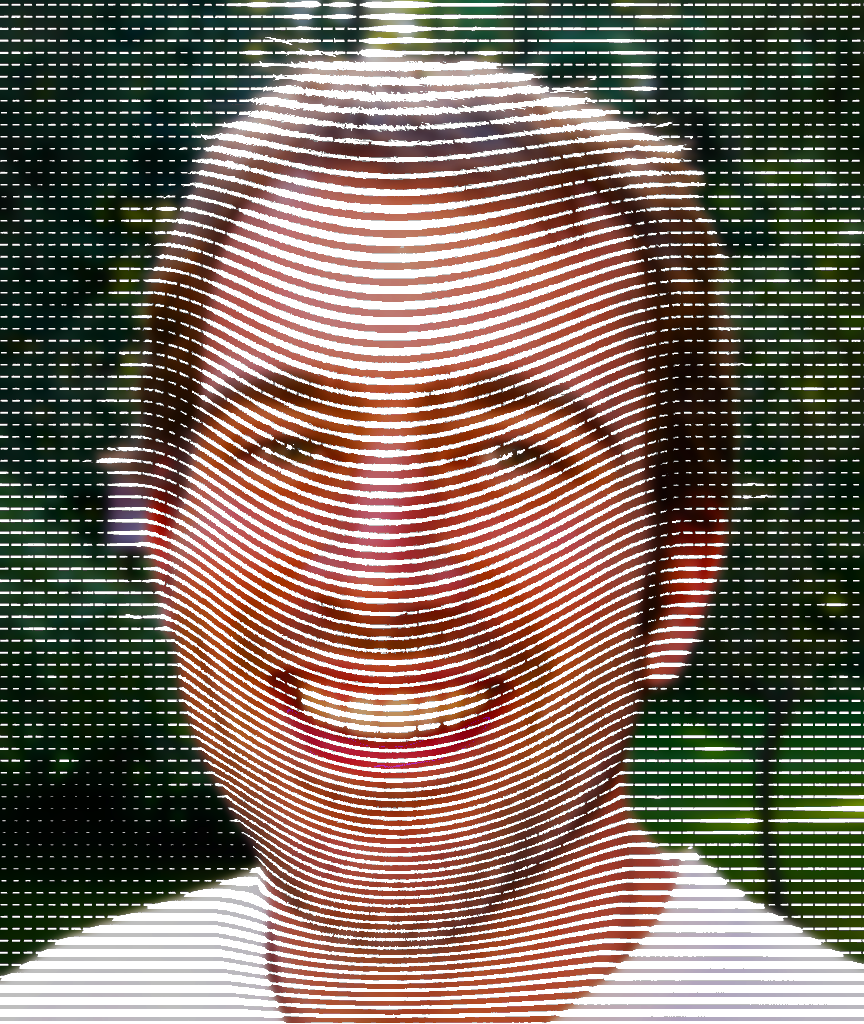}}
\subfigure[]{\includegraphics[width=0.28 \textwidth]{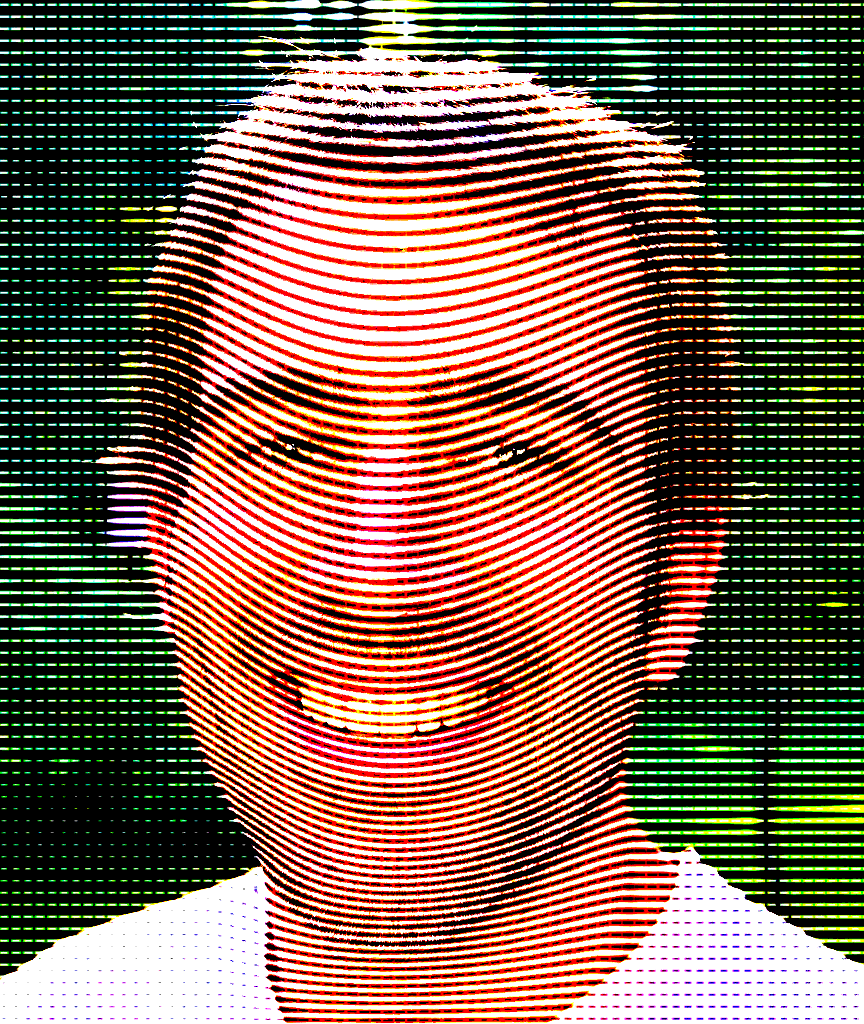}}
\subfigure[]{\includegraphics[width=0.28 \textwidth]{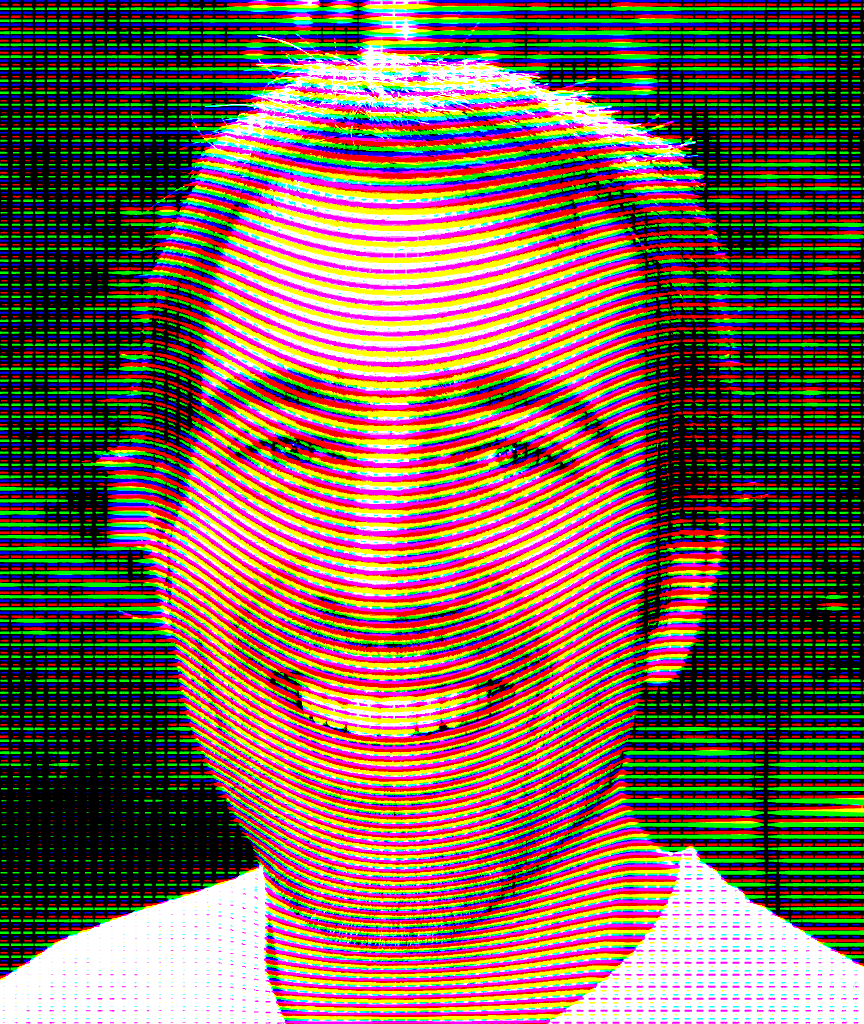}}
\caption{Coloured engraving stylisation.
a \& g) input image,
b \& h) black and white engraving,
c \& i) colour created by masking smoothed input image with black and white engraving mask,
d \& j) colour created by masking method, but processing a darker version of the image,
e \& k) colour created by applying an identical dither matrix to the three colour channels,
f \& l) colour created by applying an offset dither matrices to the three colour channels.
}
\label{colour}
\end{figure}

Engraving is traditionally a binary rendering, but following Ostromoukhov~\cite{Ostromoukhov}
we also consider colour versions of engraving.
A simple approach would be to use the black and white engraving as a mask and apply it to the colour image or a simplified version of it.
This is demonstrated in figure~\ref{colour}, in which a logical OR
of the engraving with a blurred version of the colour image produces the result in figure~\ref{colour}c\&h.
However, while the black lines of the black and white engraving have been colourised, the white lines are unchanged, and so
the overall results are lighter than the input images, and appear washed out.
This can be remedied to a degree by darkening the input image before running the engraving process,
so that consequently not only are the ORed colours darker, but also the black engraving lines used for masking are wider.
See the results in figure~\ref{colour}d\&i in which an intensity scaling of 0.75 was applied before engraving,
and subsequently the saturation of the output was scaled by a factor of 1.5 to prevent the darkening causing the engraving to look too gray.

An alternative approach is to use colour separation, and so we apply the dither matrix independently to the three RGB colour channels.
Two variants of this approach are shown in figure~\ref{colour}e\&j and figure~\ref{colour}f\&k.
The former uses the same dither matrix for the three colour channels,
which has the effect that dark (respectively light) areas in the input image tend to
appear as black (respectively white) in the colour engraving.
The second applies a vertical shift to the dither matrix by one and two thirds of its period, and
for each of the three RGB colour channels applies a different version of the dither matrix.
Shifting the dither matrix causes a separation in the engraving lines
for the colour channels, which produces a vibrant effect when viewed close-up.

\section{Conclusions}

An image-based method has been described for stylising portrait images to look like engravings
consisting of sets of parallel lines with orthogonal cross-hatchings.
This was achieved by constructing a dither matrix that when applied would generate predominantly horizontal lines,
with the addition of a lesser amount of cross-hatchings.
From detected facial landmarks a simple 3D cylindrical model for the human head was estimated,
and this enabled the deformation of the dither matrix such that the engraving lines curved around the face.

The approach is controllable, in that the proportion of cross-hatchings can be varied, as can the density of lines.
Moreover, the method can easily be applied to generate a variety of styles of colour engravings.

\bibliographystyle{abbrv}
\bibliography{engraving}

\end{document}